\RequirePackage{fix-cm}
\documentclass[smallextended]{svjour3}     
\smartqed 

\usepackage{import}
\usepackage{graphicx}
\usepackage{amssymb}

\usepackage{lineno}

\usepackage{graphicx}
\usepackage[utf8]{inputenc}
\usepackage{lmodern}
\usepackage[T1]{fontenc}
\usepackage{lscape}
\usepackage{amsmath}	
\usepackage{amssymb}
\usepackage{cases}
\usepackage{graphicx}
\usepackage{caption}
\usepackage{comment}

\usepackage{makecell}
\usepackage{multirow}
\usepackage{amsfonts}
\usepackage{float} 
\usepackage{diagbox}
\usepackage{fullpage}
\usepackage{url}
\usepackage{rotating}
\usepackage{eurosym}
\usepackage{wrapfig}
\usepackage[a4paper]{geometry}
\usepackage[nottoc]{tocbibind}
\usepackage{placeins}
\usepackage{float}
\usepackage{listings}
\usepackage{color, colortbl}
\usepackage{hyperref}
\usepackage{xcolor}
\usepackage{graphicx, caption, subcaption}
\usepackage{sidecap}
\usepackage[symbol]{footmisc}
\usepackage{caption}
\usepackage{booktabs}
\usepackage{siunitx}

\usepackage{etoolbox}
\AtBeginEnvironment{algorithm}{\SetArgSty{textrm}}  

\usepackage{sidecap}
\hypersetup{
colorlinks,
citecolor=black,
filecolor=black,
linkcolor=black,
urlcolor=black
} 

\usepackage[linesnumbered,ruled,vlined]{algorithm2e}

\hypersetup{
colorlinks,
citecolor=black,
filecolor=black,
linkcolor=black,
urlcolor=black
}

\newcommand{\argmin}{\operatornamewithlimits{argmin}}

\newcommand{\sibo}{\textcolor{black}}
\newcommand{\ming}{\textcolor{black}}

\newcommand{\bx}{\textbf{x}}
\newcommand{\xb}{\textbf{x}_b}

\newcommand{\by}{\textbf{y}}
\newcommand{\bB}{\textbf{B}}
\newcommand{\bH}{\textbf{H}}
\newcommand{\bR}{\textbf{R}}
\newcommand{\bA}{\textbf{A}}

\newcommand{\bD}{\textbf{D}}

\newcommand{\bK}{\textbf{K}}
\newcommand{\bQ}{\textbf{Q}}

\newcommand{\bC}{\textbf{C}}

\newcommand{\bW}{\textbf{W}}

\newcommand{\bI}{\textbf{I}}

\begin{document}
\title{\sibo{Observation Error Covariance Specification in Dynamical Systems for Data assimilation using Recurrent Neural Networks}}

\author{Sibo Cheng$^{1}$, Mingming Qiu$^{2,3}$ \\
        \small $^{1}$ Data Science Instituite, Department of computing, Imperial College London, UK \\
        \small $^{2}$ Institut Polytechnique de Paris, France\\
        \small $^{2}$ EDF R\&D, France
}

\institute{Sibo Cheng \at
Data Science Instituite, Department of computing, Imperial College London, UK\\
              \email{sibo.cheng@imperial.ac.uk}          
           \and
           Mingming Qiu \at
              EDF Lab Saclay, 7 Boulevard Gaspard Monge, 91120 Palaiseau, France
              \\
	      \\
              S.Cheng and M.Qiu contributed equally to this work
}

\date{\normalsize \textit{Accepted for publication in Neural computing and applications}}
\maketitle
  
\begin{abstract}
Data assimilation techniques are widely used to \sibo{predict complex dynamical systems with uncertainties}, based on \sibo{time-series observation data}. Error covariance matrices modelling is an important element in data assimilation algorithms \sibo{which can considerably impact the forecasting accuracy}. The estimation of these covariances, which usually relies on empirical assumptions and \sibo{physical constraints}, is often imprecise and computationally expensive \sibo{especially for systems of large dimension}. In this work, we propose a data-driven approach based on long short term memory (LSTM) recurrent neural networks (RNN) to improve both the accuracy and the efficiency of observation covariance specification \sibo{in data assimilation for dynamical systems}. Learning the covariance matrix from observed/simulated time-series data, the proposed approach does not require any \sibo{knowledge or assumption} about prior error distribution, unlike classical posterior tuning methods. We have compared the novel approach with two state-of-the-art covariance tuning algorithms, namely DI01 and D05, first in a Lorenz dynamical system and then in a 2D shallow water twin experiments framework with different covariance parameterization using ensemble assimilation. This novel method shows significant advantages in observation covariance specification, \sibo{assimilation} accuracy and computational efficiency.

\end{abstract}



\section{Introduction}
In order to improve the reconstruction and prediction of dynamical systems with uncertainties, data assimilation (DA) techniques, \sibo{originally developed in numerical weather prediction (NWP) \cite{F.Parrish1992} and geosciences
\cite{Carrassi2017}, are widely applied to industrial problems, such as hydrology \cite{cheng2020b}, wildfire forecasting \cite{rochoux2014towards}, drought monitoring\cite{Harisuseno2020} and nuclear engineering \cite{Gong2020}.} DA algorithms aim to find the optimal approximation (also known as the analyzed state) of the state variables (usually representing a physical field of interest, such as velocity, temperature etc.), relying on prior estimations and real-time observations, both assumed to be noisy. Due to the large dimension (often ranging from $10^6 $ to $ 10^{10}$ in NWP and geoscience problems),
prior errors are supposed to be Gaussian distributed for the sake of simplicity \cite{Asch2016}. As a consequence, the prior error distribution can be perfectly characterized by the first (mean) and the second (covariance) moment. The output of the DA algorithms is determined through some optimization function where the weight of prior simulations and observations is determined by the associated error covariance matrices, respectively named as background and observation covariances. These error covariance matrices thus provide crucial information in DA algorithms \cite{Mattern2018}, for not only the estimation of the analyzed state but also specifying posterior error distributions\cite{Eyre2013}. The prior errors represented by these matrices, especially in the case of observation errors, consist of an ensemble of different sources of noise/uncertainties, including model error, instrument noise, and representativity error \cite{dance2013,Janjic2018}.\\

In statistics, the covariance matrix of a random vector is often obtained via empirical estimation where a sufficient number of simultaneous samplings is required to avoid estimation bias \cite{Wishart1928}. Moreover, when the sampling number is inferior to the problem dimension, the estimated covariance will be rank deficient. In DA problems, the high dimensionality and lack of simultaneous data (i.e., several background or observation trajectories in the same time window) represent significant obstacles of covariance computation in data assimilation \cite{Tandeo2018}. To overcome these difficulties, we often rely on calibration (e.g., least square) methods based on some generic correlation kernels, often with homogeneous and isotropic
characteristics \cite{Fisher2003}. Balanced operators can be employed for multivariate systems \cite{Derber1989}. In terms of correlation kernels, the family of Matérn functions, including the Exponential kernel (Matérn $1/2$), the Balgovind kernel (Matérn $3/2$, also known as second order auto-regressive (SOAR) function) and the Gaussian kernel (Matérn $5/2$), is often prioritized for covariance computing owing to its smoothness and capability to capture spatial correlations in physical processes \cite{dance2013,,weston2014}. Other stationary covariance models involve, for instance, convolution formulation \cite{Gaspari1999} or diffusion-based operators \cite{Mirouze2010}, both contribute to an efficient storage of the covariance matrices. However, limited by homogeneous and isotropic assumption, it remains cumbersome to represent complex spatial correlation (often multidimensional and multivariate) using these one dimensional kernels.\\

\sibo{In this study, we develop and test a novel data-driven approach based on recurrent neural networks (RNN) to improve both the accuracy and the efficiency of observation covariance specification in dynamical data assimilation problems. The novel approach is tested and compared with two state-of-the-art covariance tuning algorithms in two different digital experiments with parametric and non-parametric covariance estimation respectively.}\\

\sibo{The paper is organized as follows. In Sect.~\ref{sec:related}, we introduce the related work for error covariance specification. The problem statement and the contribution of this paper is described in Sect.~\ref{sec:contribution}. Data assimilation techniques and the ensemble methods are introduced briefly in Sect.~\ref{sec:daaec}. We then describe traditional posterior covariance tuning algorithms DI01 and D05 in Sect.~\ref{sec:posterior tuning}. The novel LSTM-based method is introduced in Sect.~\ref{sec:lstm}, followed by the comparison in the Lorenz (Sect.~\ref{sec:lorenz}) and the shallow water twin experiments (Sect.~\ref{sec:sw}). We close the paper with a discussion in Sect.~\ref{sec:discussion}.}

\section{\sibo{Related work}}
\label{sec:related}
\sibo{To gain a clearer insight into covariance evolution}, some ensemble-based methods such as \cite{F.Parrish1992} (NMC) and \cite{Evensen1994} (EnKF), have been developed to provide a non-parametric covariance estimation. These methods depend on the propagation of an ensemble of simulated trajectories, initialized either at different forecasting time steps (NMC) or by adding some artificially set perturbations to the current state (EnKF). These methods are more appropriate for modelling the background matrix compared rather than the observation matrix. The latter, independent from the numerical simulations, can not be represented by the propagation of artificially added noises.  \sibo{The Particle-Aided Unscented Kalman Filter~\cite{lin2020self} can estimate systems with high nonlinearity with a real-time updating of the background matrix. However, the observation matrix can not be estimated directly via the Particle-Aided Unscented Kalman Filter.} In practice, the observation matrix is often set to be diagonal or spatially isotropic for the sake of simplicity (e.g. \cite{Arcucci2018}). However, it is shown in the work of \cite{dance2013} that well-specified correlated observation covariances can significantly improve the performance of DA algorithms.\\

Several methods of data-derived posterior diagnosis have also been developed based on the analysis of innovation quantities which consist of the difference between the observations and the projected background/analyzed state in the observation space. As a strong contributor to this topic, the meteorology community developed several well-known posterior diagnoses and their improved versions \cite{Desroziers01,Desroziers2005,Liu_2019} to adjust the background/observation ratio, the correlation scale length or the full covariance structure in the observation space (both the observation matrix and the projected background matrix). Some iterative processes \cite{cheng2019,kalnay2010accelerating} based on the fixed-point theory have also been proposed for error covariance tuning. Recent works of \cite{Menard2016} and \cite{Bathmann2018} have proved the convergence of so-called "Desroziers iterative methods"\cite{Desroziers2005} (also known as D05) in the ideal case. In brief, they have mathematically proved that, by using a semi positive definite matrix as an initial guess, D05 iterative method converges on the exact \footnote{Here, by the term ``exact'', we refer to the covariance truly corresponding to the remaining errors present in the observation space} time-invariant (at least over a sufficiently long time period) observation error covariance when the background matrix and the transformation operator (which maps the state variables to real-time observations) are perfectly known \textit{a priori}. On the other hand, it is also mentioned by \cite{Bathmann2018} that a regularization step is necessary in practice for applying D05 and the convergence of the regularized iterations remains an open question \cite{Bathmann2018,cheng2020b}.
 To deal with time-varying systems, lag-innovation statistics are used for error covariance estimation \cite{DaleyMWV1992}. The essential idea is to build a secondary Kalman-filtering process for adjusting error covariances using time-shifted innovation vectors. For more details of the innovation-based methods, we refer to the overview of \cite{Tandeo2018} which also covers some other estimation methods, such as the family of likelihood-based approaches and expectation-maximum(EM) methods.\\

\section{\sibo{Problem statement and contribution}}
\label{sec:contribution}
\sibo{Our work lies in a similar condition of \cite{Desroziers2005} and \cite{Bathmann2018} where both the state forward model and the transformation operator are presumed to be well known. As the main difficulty concerns the non-synchronous time-variant observations in dynamical systems (which prevents empirical estimation),  in this work we propose the use of recurrent neural networks (RNN) \cite{Rumelhart1987} for the specification of the observation matrix across the underlying dynamics of the observed quantities.} \sibo{RNN has been widely adapted for the prediction/reconstruction of dynamical systems, especially in natural language processing (NLP) \cite{Yin2017ComparativeSO} and image/video processing \cite{Sarabu2021} due to its convincing capacity of dealing with time series.}  More recently, RNN has also made their way to other engineering fields such as biomedical applications and computational fluid mechanics \cite{Suraj2020}. In general, the combination of deep learning and data assimilation methods \cite{Arcucci2021,Geer2021} has been widely adapted and analyzed in a variety of industrial applications, including air pollution \cite{CASAS2020} and ocean-atmosphere modelling \cite{Brajard2021}. A convolutional neural network (CNN) for covariance estimation has also been suggested in the work of \cite{Liu2018}. In this study, we propose a novel methodology for LSTM-based covariance estimation which can be easily integrated into any DA schema for dynamical systems.
Here, we first construct a set of training covariance matrices, being either parametric or non-parametric, within a certain range defined \textit{a priori}. For each matrix in the training set, we then simulate a dynamic trajectory of the observation vector relying on the knowledge of the forward model where the noises at each time step are generated following a centered Gaussian distribution characterized by the error covariance. These trajectories are later used as input variables to train the long-short-term-memory (LSTM) RNN regression model
where the time invariant observation matrices stand for the learning target. For the online evaluation, only the historical observation data is needed to predict the error covariances. Compared to traditional posterior tuning methods \cite{Desroziers2005,Dreano2017} which require several implementations of DA algorithms, the proposed machine learning (ML) method can be much more computationally efficient for real-time covariance estimation. Moreover, no prior knowledge concerning either the background or the observation matrix is necessary for the proposed ML approach unlike most of the traditional methods. For example, DI01~\cite{Desroziers01} requires precise knowledge of correlation structures for both background and observation matrices while D05~\cite{Desroziers2005} make use of the perfect knowledge of the background covariance.\\

In order to make a comprehensive comparison with traditional methods, two different twin experiment frameworks are implemented in this paper, using respectively the Lorenz96 and the 2D shallow water models. The Lorenz system, characterized by only three state variables, is associated with a non-parametric covariance modelling  while we use an isotropic correlation kernel to parameterize the observation matrix in the shallow water dynamics. In both cases, we compare the performance of the proposed LSTM-based method against the 
state-of-the-art tuning approaches D05 and DI01 in terms of both the covariance specification and the posterior DA accuracy. An ensemble DA schema is used for estimating the time variant background matrix for each of these methods.\\

\section{Data assimilation}
\label{sec:daaec}

\subsection{Principle of data assimilation}
\label{sec:eda}

The objective of data assimilation algorithms is to approach the estimation of system's states $\textbf{x}$ to its true values $\textbf{x}_\textrm{true}$, also known as the true state, by taking advantage of two sources of information: the prior estimation or forecast $\textbf{x}_b$, which is also called the background state, and the measurement or observation $\textbf{y}$. DA algorithms aim to find an optimally weighted compromise between $\textbf{x}_b$ and $\textbf{y}$ by minimizing the lost function $J$ defined as: 
\begin{align}
    J(\textbf{x})&=\frac{1}{2}(\textbf{x}-\textbf{x}_b)^T\textbf{B}^{-1}(\textbf{x}-\textbf{x}_b) + \frac{1}{2}(\textbf{y}-\mathcal{H}(\textbf{x}))^T \textbf{R}^{-1} (\textbf{y}-\mathcal{H}(\textbf{x})) \label{eq_3dvar}\\
   &=\frac{1}{2}||\textbf{x}-\textbf{x}_b||^2_{\textbf{B}^{-1}}+\frac{1}{2}||\textbf{y}-\mathcal{H}(\textbf{x})||^2_{\textbf{R}^{-1}},
\end{align}
 where  $\mathcal{H}$ denotes the transformation operator from the state space to observation space. $\textbf{B}$ and $\textbf{R}$ are respectively the background and the observation error covariance matrices, i.e.
 \begin{align}
     \textbf{B} = \textrm{Cov}(\epsilon_b, \epsilon_b), \quad
     \textbf{R} = \textrm{Cov}(\epsilon_y, \epsilon_y),
 \end{align}
 where
  \begin{align}
     \epsilon_b = \textbf{x}_b - \textbf{x}_\textrm{true}, \quad
     \epsilon_y = \mathcal{H}(\textbf{x}_\textrm{true})-\textbf{y}.
 \end{align}

Errors $\epsilon_b, \epsilon_y$ are supposed to be centered Gaussian following:
 \begin{align}
     \epsilon_b \sim \mathcal{N} (0, \textbf{B}), \quad
     \epsilon_y \sim \mathcal{N} (0, \textbf{R}).
 \end{align}
 
In Eq.~\ref{eq_3dvar}, the left side strives for incorporating the prior information $\textbf{x}_b$, and the right side penalizes the difference between the observation $\textbf{y}$ and the state variables after having been mapped to the observation space $\mathcal{H}(\textbf{x})$. Both terms are weighted by the corresponding inverse of error covariance matrix ($\textbf{B}^{-1}$, $\textbf{R}^{-1}$) to reflect confidences for each of them.\\
 
 The optimization problem of Eq.~\ref{eq_3dvar}, so called three-dimensional variational (3D-Var) formulation, is a general representation of variational assimilation which does not take into account model errors. The output of Eq.~\ref{eq_3dvar} is denoted as $\bx_a$, i.e.
  \begin{align}
    \bx_a = \underset{\bx}{\argmin} \Big(J(\textbf{x})\Big). \label{eq:argmin}
 \end{align}

If $\mathcal{H}$ is the linear observation operator represented by $\bH$, Eq.~\ref{eq:argmin} can be solved via BLUE (Best Linearized Unbiased Estimator) formulation:
   \begin{align}
    \textbf{x}_a &= \textbf{x}_b+\textbf{K}(\textbf{y}-\textbf{H} \textbf{x}_b), \\
    \bA &= (\textbf{I}-\textbf{K}\textbf{H})\textbf{B}, \label{eq:BLUE}
 \end{align}
 where $\textbf{A} = \textrm{Cov}(\textbf{x}_a-\textbf{x}_\textrm{true})$ is the analyzed error covariance, and $\textbf{K}$ is the Kalman gain matrix described by
 \begin{equation}
	\textbf{K}=\textbf{B} \textbf{H}^T (\textbf{H} \textbf{B} \textbf{H}^T+\textbf{R})^{-1}.
	\label{eq:Kgain_BLUE}
\end{equation}

In the rest of this paper, we define $\bH$ as a linear transformation operator. Nevertheless, it is usually more challenging to  find the minimum of Eq.~\ref{eq_3dvar} when $\mathcal{H}$ is non-linear, even more challenging when states are high-dimensional. The resolution for the minimization often involves gradient descent algorithms (such as "L-BFGS-B" \cite{Fulton2000} or adjoint-based \cite{Cioaca2014} numerical techniques. 

DA algorithms could be applied to dynamical systems thanks to sequential applications expressed by a transition operator $\mathcal{M}_{t^k \rightarrow t^{k+1}}$ (from discrete time $t^{k}$ to $t^{k+1}$), where
\begin{align}
    \textbf{x}_{t^{k+1}} = \mathcal{M}_{t^k \rightarrow t^{k+1}} (\textbf{x}_{t^k}).
\end{align}

$\textbf{x}_{b,t^{k+1}}$ thus depends on the knowledge of $\mathcal{M}_{t^k \rightarrow t^{k+1}}$ and the DA correcting state $ \textbf{x}_{a,t^k}$, i.e.,
\begin{align}
    \textbf{x}_{b,t^{k+1}} = \mathcal{M}_{t^{k} \rightarrow t^{k+1}} (\textbf{x}_{a,t^{k} })
    \label{eq:x_b_t_acquiring}
\end{align}

Obviously, the more accurate $\textbf{x}_{a,t^{k}}$is, the more reliable $\textbf{x}_{b,t^{k+1}}$ would be.\\

To leverage the information embedded in the background state and observations, covariance matrices modeling is a pivotal point in DA as they influence not only how prior errors spread but may also change the DA results \cite{cheng2019}.
\subsection{Ensemble methods}
\label{sec:EnDA}
Ensemble data assimilation (EnDA) \cite{Bannister2017,bazargan2019} methods have shown a strong performance in dealing with non-linear chaotic DA systems
by creating an ensemble with size $M$ of the system state depicted as $\{\textbf{x}_{t^{k}}^{(i)}|1\leq i \leq M\}$. The latter is used to represent both the prior and the posterior probability distribution of the state variables. The system states of the ensemble evolve under $\mathcal{M}_{t^{k} \rightarrow t^{k+1}}$ and DA is applied to each of these ensemble states at every assimilation windows. Furthermore, instead of evolving the system to obtain the $\textbf{B}$ matrix, which is a time and computationally expensive process when a large number of states is available, $\textbf{B}$ is estimated as a sample covariance:

\begin{align}
\label{eq:ensemble_b}
     \textbf{B}_{b,t^{k}} \approx \frac{1}{M-1}\sum_{i=1}^{M}(\textbf{x}_{b,t^{k}}^{(i)}-\overline{\textbf{x}}_{b,t^{k}})(\textbf{x}_{b,t^{k}}^{(i)}-\overline{\textbf{x}}_{b,t^{k}})^{T},
\end{align}
where $\overline{\textbf{x}}_{b,t^{k}}=\frac{1}{M-1} \sum_{i=1}^{M}\textbf{x}_{b,t^{k}}^{(i)}$, and the estimation becomes more reliable with the increases of $M$. For applications in this study, EnDA, with a sufficiently large number of examples, is used to estimate $\textbf{x}_{b,t^{k}}$ and $\textbf{B}_{b,t^{k}}$ so that we can focus on the comparison of $\bR$ matrix modelings.

\subsection{Observation error covariances specification}
 For the estimation of $\textbf{R}$, under the assumption that the system model is stationary, a wide variety of methods have been explored, for example, the DI01 \cite{Desroziers01} method which adjusts accordingly the ratio between $Tr(\bB)$ and $Tr(\bR)$ and the D05 \cite{Desroziers2005} approach which estimates the full observation space iteratively.  However, these methods, based on posterior innovation quantities (i.e., $\by-\mathcal{H}(\bx_a)$) which requires several applications of DA algorithms, can be computationally expensive. Moreover,  these tuning methods, especially the D05 which estimates the full matrix, are not suitable for different matrix parameterizations.
In this paper, working with time-series observation data, we use LSTM  to predict the corresponding $\textbf{R}$ matrix under similar assumptions of DI01 and D05. The two classical methods, introduced in Sect.~\ref{sec:posterior tuning}, are implemented to compare the results with the proposed machine learning approach.

\section{Posterior covariance tuning algorithms}
\label{sec:posterior tuning}
\subsection{Desroziers \& Ivanov (DI01) tuning algorithm}
Because $\bB$ and $\bR$ determine the weight of background and observation information in the loss function ( Eq.~\ref{eq_3dvar}), the knowledge of $\textrm{Tr}(\bB)$ and $\textrm{Tr}(\bR)$ is crucial to DA accuracy.
DI01 \cite{Desroziers01} tuning algorithm, relying on the diagnosis of innovation quantities, has been widely adopted in meteorology \cite{Michel2014,Menard2016} and geoscience \cite{Hoffman2013}. Consecutive works have been carried out to improve its performance and feasibility in problems of large dimension \cite{Chapnik2006}. Without modifying error correlation structures, DI01 adjusts the prior error amplitudes by applying an iterative fixed-point procedure. 

As demonstrated in \cite{Talagrand99,Desroziers01}, when $\bB$ and $\bR$ are perfectly specified,
\begin{align}
	\mathbb{E}\left [J_b(\textbf{x}_a) \right ] & = \frac{1}{2} \mathbb{E}\left [(\textbf{x}_a-\textbf{x}_b)^T \bB^{-1}(\textbf{x}_a-\textbf{x}_b) \right ] \notag \\
	& =\frac{1}{2} \textrm{Tr}(\bK \bH), \label{eq:Jb} 
\end{align} 
\begin{align}
	\mathbb{E} \left [J_o(\textbf{x}_a)  \right ] & = \frac{1}{2} \mathbb{E}\left [(\textbf{y}-\bH\textbf{x}_b)^T \bR^{-1}(\textbf{y}- \bH\textbf{x}_b) \right ] \notag \\
	& =\frac{1}{2} \textrm{Tr}(\mathbf{I}-\bH \bK), \label{eq:Jo1} 
\end{align}
 where $\bH$ is a linearized observation operator. Based on Eq.~\ref{eq:Jb} and \ref{eq:Jo1} it is possible to iteratively correct the magnitudes of $\bB$ and $\bR$, following
 \begin{equation}
	\bB_{q+1}=s_{b,q} \bB_q, \quad \bR_{q+1}=s_{o,q} \bR_q,
	\label{eq:B_R_iterated_update}
\end{equation}
using the two indicators
\begin{equation}
	s_{b,q}=\frac{2J_b(\textbf{x}_a)}{\textrm{Tr}(\bK_q \bH)},
\end{equation}
\begin{equation}
	s_{o,q}=\frac{2J_o(\textbf{x}_a)}{\textrm{Tr}(\mathbf{I}-\bH \bK_q)}, \label{eq:DI01}
\end{equation}
where $q$ is the current iteration.

Acting as scaling coefficients, the sequences $\{ s_{b,q}\}$ and $\{ s_{o,q}\}$ modify the error variance magnitude in the iterative process. It is worth reminding that both the analyzed state $\textbf{x}_a$ and the gain matrix $\bK_q$ are obtained using $\bB_q$, $\bR_q$ which depend on $s_{b,q}$ and $s_{o,q}$. When the correlation patterns of both $\bB$ and $\bR$ are well known, DI01 is equivalent to a maximum-likelihood parameter tuning, as pointed out in \cite{Chapnik2006,Menard2016}.

Unlike other posterior covariance diagnosis/computations, such as \cite{Desroziers2005,cheng2019}, the estimation of the full matrices is not needed in DI01. Instead, only the estimation of two scalar values ($J_b,J_o$) is required, which significantly reduce the computational cost. As a consequence, this method could be more appropriate for online covariance tuning.

\subsection{Desroziers iterative method (D05) in the observation space}

The Desroziers diagnosis (D05)\cite{Desroziers2005}, subject to prior and posterior state-observation residuals has been widely applied in engineering problems, including numerical weather prediction\cite{Menard2016} and hydrology \cite{cheng2020b}. The work of \cite{Desroziers2005} proved that when $\textbf{B}$ and $\textbf{R}$ are well known a priori, the expectation
of the analysis state should satisfy:
\begin{align}
    \mathbb{E}\Big([\textbf{y}-\mathcal{H}(\bx_a)][\textbf{y}-\mathcal{H}(\bx_b)]^T\Big) &= \bR. \label{diag_1}
\end{align}
The difference between the left side and the right side of
Eq.\ref{diag_1}, 
\begin{align}
    || \textbf{R} -  \mathbb{E}\big([\textbf{y}-\mathcal{H}(\bx_a)][\textbf{y}-\mathcal{H}(\bx_b)]^T \big)||_F, \label{indi_R}
\end{align}
can be used as a validation indicator of the $\textbf{R}$ matrix estimation where
$||.||_F$ denotes the Frobenius norm. Applying this method, time variant observation/background data can contribute to the
estimation of the $\bR$ matrix because the expectation in Eq.~\ref{diag_1} could be evaluated using residuals at different time steps. When the $\bB$ matrix is well known, an iterative process has been introduced to estimate the $\bR$ matrix:
\begin{align}
    \bR_{n+1} = \mathbb{E}\Big([\textbf{y}-\mathcal{H}(\bx_{a,q})][\textbf{y}-\mathcal{H}(\bx_b)]^T \Big),  \label{eq:it_D05}
\end{align}
based on the fixed-point theory \cite{Bathmann2018}. The current analysis state $\bx_{a,q}$ is obtained using the
specification of $\bR_q$ while $\xb, \bB, \by$
remains invariant. As proved in \cite{Menard2016} and \cite{Bathmann2018}, under the assumption of sufficient observation data and well known
$\bB$ matrix, the iterative process of Eq.\ref{eq:it_D05} converges to the exact
observation error covariance. However, as shown in \cite{Bathmann2018},
 the intermediate matrices $\bR_q$ could be non-symmetric and possibly contain
negative or complex eigenvalues, which is cumbersome for DA algorithms to deal with.\\

In practice a posterior regularization at each
iteration step is often required to ensure the positive definiteness of $\bR_q$ \cite{Bathmann2018} where the first step of the
regularization could be symmetrizing the estimated $\bR_q$ matrix, i.e.,
\begin{align}
    \bR_q \longleftarrow \frac{1}{2} (\bR_q + \bR^T_q). \label{eq:symR}
\end{align}
 The spectrum of
$\bR_q$ now contains only real numbers but they are not necessarily positive. The hybrid method\cite{Carrassi2017} is a standard
approach in ensemble-based DA methods to ensure the positive definiteness, which consists of combining a prior defined
covariance matrix  $\textbf{C}$ with the one obtained from empirical estimation. We thus obtain the formulation of the
regularized observation matrix $\bR_\textrm{r,n}$:
\begin{align}
    \bR_\textrm{n} \longleftarrow (1-\mu)\bR_q + \mu \textbf{C}, \label{eq:hybrid}
\end{align}
following Eq.\ref{eq:symR} with $\mu \in (0,1)$. The matrix $\textbf{C}$ is
often set as a diagonal matrix since it helps to enhance the matrix conditioning. In this work, we choose to set
\begin{align}
   \mu = 0.2 \quad \textrm{and} \quad \bC = \frac {Tr(\bR_q) \times \bI}{\textrm{dim}(\by)},
\end{align}
so that $Tr(\bR_q)$ will not be modified due to the post regularization.
As mentioned in the discussion of \cite{Bathmann2018,cheng2020b}, the convergence
of regularized observation matrices remains an open question. Therefore, a small iteration number is often assigned for D05 tuning in industrial problems (e.g., $q=2$ in \cite{Migliorini2013} and \cite{cheng2020b}). Since the right side of Eq.~\ref{eq:it_D05} can be estimated using residual quantities at different time steps, D05 is often used to deal with time series observation data (e.g., \cite{Migliorini2013,cheng2020b}) when assuming the $\bR$ matrix is time-invariant. 

\section{LSTM for error covariance estimation}
\label{sec:lstm}

\subsection{Introduction of RNN and LSTM}
\begin{figure}[!ht]
  \makebox[\textwidth][c]{\includegraphics[width=1\textwidth]{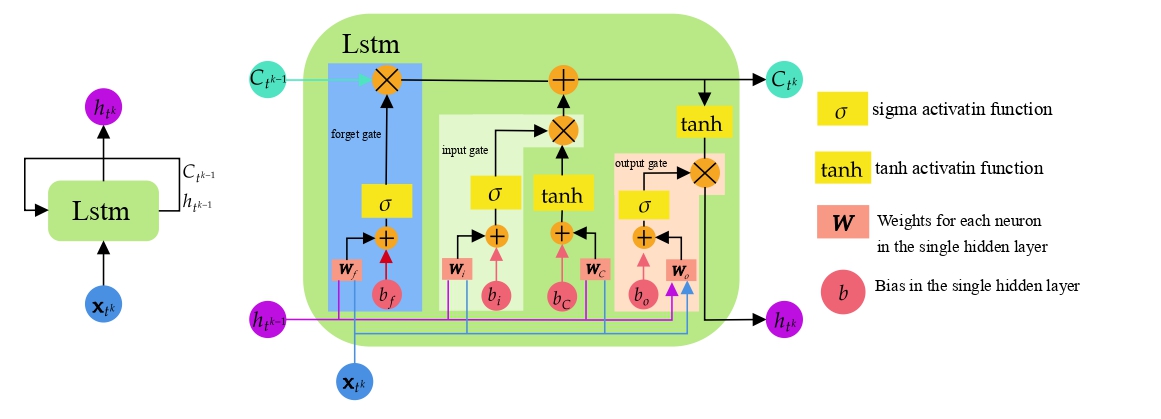}}%
  \caption{LSTM diagrams}
  \label{fig:lstm}
\end{figure}

LSTM, first introduced in \cite{hochreiter1997long}, is a kind of RNN \cite{Rumelhart1987}, capable of solving long term dependency problems \cite{bengio1994learning} that traditional RNN could not handle. As with other recurrent neural networks, LSTM has a chain-like structure. This structure is created by repeating the same module shown on the left side in Fig.~\ref{fig:lstm}. In addition, the repeating module comprises four neural networks instead of only one. The specific structure of the repeating module is on the right side in Fig.~\ref{fig:lstm}. 

An essentail part of LSTM is the cell state $\mathbf{C}_{t^{k-1}}$ which is the long-term memory storing information about past behaviors. LSTM uses three gates with each composed out of a sigmoid layer neural network (single layer neural network with sigmoid activation function at the output layer) and a pointwise multiplication operation, to protect and control information of the cell state as shown in Fig.~\ref{fig:lstm}. 

The first gate is the forget gate following: 
\begin{align}
\label{eq:forget_gate}
  \mathbf{f}_{t^k}=\sigma(\mathbf{W}_f\cdot[\mathbf{h}_{t^{k-1}},\mathbf{x}_{t^k}]+b_f),
\end{align}

where the recurrent variable $\mathbf{h}_{t^{k-1}}$ summarizing all the information about past behaviors, $\mathbf{x}_{t^k}$ resuming information about current behaviors, and $\mathbf{W}_f$ and $b_f$ are weights and bias respectively parameterizing the sigma layer neural network. The forget gate decides what kind of information is going to be ignored in $\mathbf{C}_{t^{k-1}}$. 

The second gate is the input gate, and it determines which new information is added into $\mathbf{C}_{t^{k-1}}$. This new information $\tilde{\mathbf{C}}_{t^{k}}$, as conforming to
\begin{align}
\tilde{\mathbf{C}}_{t^{k}}=tanh(\mathbf{W}_c\cdot[\mathbf{h}_{t^{k-1}},\mathbf{x}_{t^k}]+b_c),
\label{eq:tanh_layer}
\end{align}
is attained by passing $\mathbf{h}_{t^{k-1}}$ and $\mathbf{x}_{t^k}$ to a $tanh$ layer neural network (single layer neural network with tanh activation function at the output layer) with parameters $\mathbf{W}_c$ and $b_c$.

$\tilde{\mathbf{C}}_{t^{k}}$ is then multiplied by weight coefficients $\mathbf{i}_{t^{k}}$ which is acquired by the input gate, i.e.,
\begin{align}
\mathbf{i}_{t^{k}}=\sigma(\mathbf{W}_i\cdot[\mathbf{h}_{t^{k-1}},\mathbf{x}_{t^{k}}]+b_i).
\label{eq:input_gate}
\end{align}
$\mathbf{i}_{t^{k}}$ is applied to decide which new information would be employed to update $\mathbf{C}_{t^{k-1}}$.

Hence, the state cell $\mathbf{C}_{t^{k}}$ at the current time step $t^k$ can be attained using
\begin{align}
\label{eq:cell_state}
\mathbf{C}_{t^{k}}=\mathbf{f}_{t^k}\odot \mathbf{C}_{t^{k-1}}+\mathbf{i}_{t^{k}}\odot \tilde{\mathbf{C}}_{t^{k}}.
\end{align}

Finally, the acquisition of $\mathbf{h}_{t^{k}}$ requires the participation of the output gate and a tanh activation function: first, the tanh activation function $tanh$ is used to create a cell state candidate information $tanh(\mathbf{C}_{t^{k}})$. $tanh(\mathbf{C}_{t^{k}})$ is then multiplied by some weight coefficients following
\begin{align}
\label{eq:recurrent_output}
\mathbf{h}_{t^{k}}=\mathbf{o}_{t^{k}}\odot tanh(\mathbf{C}_{t^{k}}),
\end{align}
to decide which information of $tanh(\mathbf{C}_{t^{k}})$ would contribute to the obtainment of $\mathbf{h}_{t^{k}}$.  
Among them, $\mathbf{o}_{t^{k}}$ is generated by the output gate with neural network parameters $\mathbf{W}_o$ and $b_o$, i.e.,
\begin{align}
\label{eq:output_gate}
\mathbf{o}_{t^{k}}=\sigma(\mathbf{W}_o[\mathbf{h}_{t^{k-1}},\mathbf{x}_{t^{k}}]+b_o).
\end{align}

\subsection{LSTM for $\bR$ matrix estimation using time series observation data}
The tuning methods presented in Sect.~\ref{sec:posterior tuning} have been applied in various engineering applications with significant improvement of covariance specification and DA accuracy. \ming{However, these methods which require several applications of DA algorithms can be computationally expensive for high-dimensional problems}. Another important drawback stands for the requirement of precise knowledge on either the correlation patterns of $\bB$ and $\bR$ (for DI01) or the full $\bB$ matrix (for D05). 
In this study, we aim to build a data-driven surrogate model for efficient online $\bR$ matrix specification using LSTM. Unlike DI01 or D05, no specific knowledge about the error covariances or the state/observation dynamical systems other than the transformation operator $\mathcal{H}$ and the forward model $\mathcal{M}_{t^k \rightarrow t^{k+1}}$  (which is also indispensable for standard DA algorithms including variational methods and Kalman-filters) is required. \\

Based on an initial state \ming{$\bx_{b,t^{0}}^{[\textrm{iter}]}=\bx_{g,t^{0}}^{[\textrm{iter}]}$, where $[\textrm{iter}]$ is the indication of some certain application and $\bx_{g,t^{0}}^{[\textrm{iter}]}$ suggests the initial generated state set of this application}, our main idea is to build a training set for the specific problem, including predefined time-invariant observation matrices $\{\textbf{R}^{[\textrm{iter}]}\}$ within certain range and generated dynamical observation vector $\{ \mathbf{y}_{t^k}^{[\textrm{iter}]} \}$.
Setting the dynamical observations $\{ \mathbf{y}_{t^k}^{[\textrm{iter}]} \}$ as the system input and the $\bR$ matrices as output, LSTM networks are used to learn the error distribution across the underlying observation dynamics. More precisely, a real function $g^{\bR}(.): \Phi_\textbf{R} \longrightarrow \mathbb{R}^{m \times m}$ is predefined where $\Phi_\textbf{R}$ is an empirically estimated real space which defines the range of a set of parameters, such as marginal error variance, correlation scale length \cite{Gaspari1999} for computing the $\bR$ matrices. The generated observation matrices $\{\textbf{R}^{[\textrm{iter}]}\}$ are set to be symmetric positive definite (SPD) thanks to the function $g^{\bR}(.)$. Both $g^{\bR}(.)$ and $\Phi_\textbf{R}$ vary for different applications.\\

Generated states $\{ \textbf{x}^{[\textrm{iter}]}_{g,t^{k}}\}$, $t^{k}\in \{0,\cdots t^T\}$ with $t^T$ depicting the index of the final time step, could be attained by evolving the system respecting $\mathcal{M}_{t^k \rightarrow t^{k+1}}$:
\begin{align}
    \ming{\{\textbf{x}^{[\textrm{iter}]}_{g,t^{k+1}}\} = \mathcal{M}_{t^k \rightarrow t^{k+1}}(\{\textbf{x}^{[\textrm{iter}]}_{g,t^{k}}\})}.
\end{align} 
The observation $\{ \textbf{y}^{[\textrm{iter}]}_{t^{k}} \}$, $t^{k}\in \{0,\cdots t^T\}$, is gained by mapping  $\{ \textbf{x}^{[\textrm{iter}]}_{g,t^{k}}\}$ through $\mathcal{H}$ and at the same time combining with random Gaussian noises:
 \begin{align}
     \ming{\{\textbf{y}^{[\textrm{iter}]}_{t^{k}}\} = \mathcal{H}(\{ \textbf{x}^{[\textrm{iter}]}_{g,t^{k}}\}) + \{\epsilon^{[\textrm{iter}]}_{g,t^{k}}\}} \quad & \textrm{for} \quad  t^{k}\in \{0,\cdots t^T\} \quad \textrm{and} \quad \textrm{iter} = 1...N, \notag \\
     \textrm{where } \quad \ming{\{\epsilon^{[\textrm{iter}]}_{g,t^{k}}\}} &\sim \ming{\mathcal{N}(0,\{\textbf{R}^{[\textrm{iter}]}\}}).
 \end{align}

\begin{algorithm}[ht]
\SetAlgoLined
Parameters: \\
Number of simulated examples: $N$\\
Total time steps: $t^{T}$\\
Covariance parameters range: $\Phi_\textbf{R}$\\
System transition operator: $\mathcal{M}_{t^k\rightarrow t^{k+1}}$\\
Observation operator: $\mathcal{H}$ \\
\BlankLine
Inputs:\\
\ming{Initial state value: $\bx_{t^0}$}  \\
Space of parameters: $\Phi_\textbf{R}$ \\
R matrix generation function: $g^{\bR}(\Phi_\textbf{R})$ \\
\BlankLine
Algorithm:\\
\For{$\textrm{iter}\leq N$}{
    $\textbf{R}^{[\textrm{iter}]}=g^{\bR}(\Phi_\textbf{R}^{[\textrm{iter}]})$\\
    
    \For{$t^k\leq t^T$}{
        \ming{$\bx_{t^k}^{[\textrm{iter}]}=\mathcal{M}_{t^k\rightarrow t^{k+1}}(\textbf{x}_{t^{k-1}}^{[\textrm{iter}]})$}\\
         $\textbf{y}_{t^k}^{[\textrm{iter}]}=\mathcal{H}(\textbf{x}_{t^k}^{[\textrm{iter}]}) + \epsilon^{[\textrm{iter}]}_{g,t^{k}}$, with $\epsilon^{[\textrm{iter}]}_{g,t^{k}} \sim \mathcal{N}(0,\textbf{R}^{[\textrm{iter}]})$
    }
}

$return$ $\{\textbf{R}^{[\textrm{iter}]}\}$, $\{\textbf{y}^{[\textrm{iter}]}\}$, ${\textrm{iter} = 0..N}$

\caption{Simulated observations $\textbf{y}$ generating process}
\label{ALgo:observationGeneration}
\end{algorithm}

\begin{figure}[H]
  \centering
  \includegraphics[width=6.3 in]{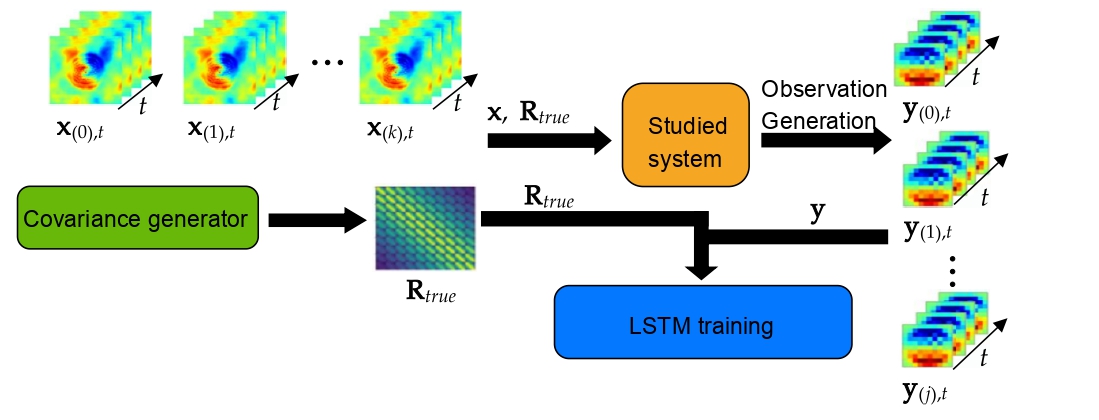}
    \caption{Offline training schema of the proposed method, including covariance generation, observation generation and LSTM training}
    \label{fig:data_generation_training}
\end{figure}

After having generated $\{\textbf{R}^{[\textrm{iter}]}\}$ and $\{ \mathbf{y}_{t^k}^{[\textrm{iter}]} \}$ as illustrated in Algo.~\ref{ALgo:observationGeneration},
a LSTM network is then trained to predict $\{\textbf{R}^{[\textrm{iter}]}\}$ knowing $\{ \mathbf{y}_{t^k}^{[\textrm{iter}]} \}$. The general process can be demonstrated in Fig.~\ref{fig:data_generation_training} where \ming{for each application,} $\by^{[\textrm{iter}]}$ is simulated by evolving the system knowing the system dynamics and the observation error covariance generator, and then $\by^{[\textrm{iter}]}$ and $\bR^{[\textrm{iter}]}$ are applied to train LSTM so that LSTM can predict $\bR^{[\textrm{iter}]}$ when receiving unseen $\by^{[\textrm{iter}]}$.

\begin{figure}
  \makebox[\textwidth][c]{\includegraphics[width=0.8\textwidth]{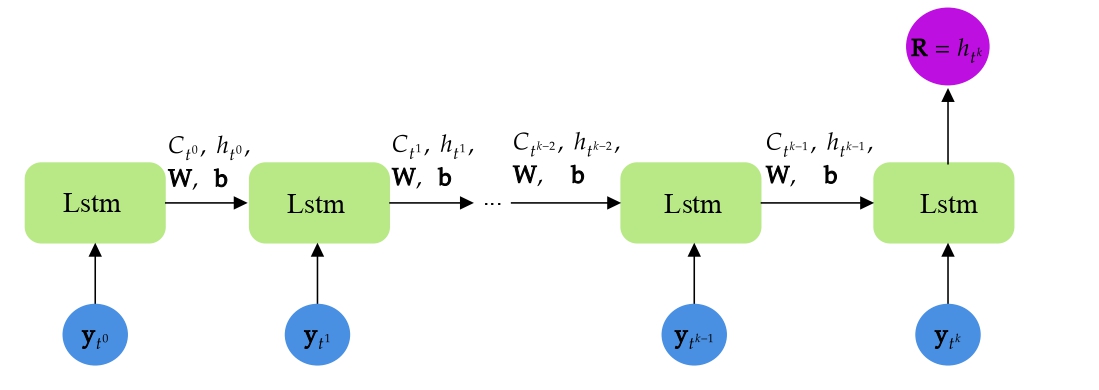}}%
  \caption{Many to one LSTM training and $\textbf{R}$ prediction process}
  \label{fig:many_to_one_lstm}
\end{figure}

Following the principle of many to one LSTM in Fig.~\ref{fig:many_to_one_lstm}, the input features of LSTM consist of observations $\by^{[\textrm{iter}]}=\{\by_{t^0}^{[\textrm{iter}]},\by_{t^1}^{[\textrm{iter}]},\cdots,\by_{t^k}^{[\textrm{iter}]},\cdots,\by_{t^T}^{[\textrm{iter}]}\}$, while the output is $\bR^{[\textrm{iter}]}$ matrix. Different from classical covariance tuning algorithms, the LSTM network only makes use of historical observation data, requiring neither the background states nor the error covariance matrix. The advantage of using LSTM is more salient when the observation dimension is large, for example, millions or even billions, while such dimension is not uncommon in real-world applications \cite{Asch2016,Carrassi2017}.

To estimate $\bR$, LSTM is first trained to learn: -- either related variables which can be used to constitute the symmetry observation error covariance matrix (i.e., input variables of the $g^{\bR}(.)$ function) in a parametric modeling; -- or elements of the $\bR$ matrix (e.g., variables in the upper triangle and those in the diagonal of the covariance matrix) in a non-parametric modeling. The whole process for $\bR$ estimation using LSTM is described in Algorithm \ref{ALgo:LstmForR}.

\begin{algorithm}
\SetAlgoLined
Parameters:\\
Number of simulated examples: $N$\\
Total time steps: $t^T$\\
Portion between validation and total examples: $\rho$ \\
LSTM input neurons (length): $N_{\textrm{in}}$\\
LSTM output neurons (length): $N_{\textrm{out}}$\\
Epoch size: $N_{\textrm{epoch}}$\\
Batch size: $L_{\textrm{batch}}$\\
Patience epoch size: $N_{\textrm{patience}\underline{\hspace{0.2cm}}\textrm{epoch}}$\\
\BlankLine
Inputs:\\
Observations: $\by_{t^k}^{[\textrm{iter}]}$ for $t^k\in\{0,\cdots,t^T\}$, $\textrm{iter}\in \{0,\cdots,N\}$ \\
Observation error covariance matrix: $\bR_{\textrm{true}}^{[\textrm{iter}]}$ for $\textrm{iter}\in\{0,\cdots,N\}$\\
Designed LSTM function: $f^L$\\
Weight parameters of neural network in Lstm: $\bW$\\
\BlankLine
Algorithm:\\
\While{$n_{\textrm{epoch}}\leq N_{\textrm{epoch}}$}{

\For{$l_{\textrm{batch}} \leq (N\cdot(1-\rho))/L_{\textrm{batch}}$}
{
    \For{$\textrm{iter}\leq L_{\textrm{batch}}$}
    {
        $(\bR^{[\textrm{iter}]})\leftarrow f^L(\by_{t^0}^{[\textrm{iter}]},\cdots,\by_{t^T}^{[\textrm{iter}]};N_{\textrm{in}},N_{\textrm{out}},\bW )$\\
        
        $\textrm{train}\underline{\hspace{0.2cm}}\textrm{loss}=({\bR^{[\textrm{iter}]}-\bR_{\textrm{true}}^{[\textrm{iter}]})}^2$\\
        
        $\bW\leftarrow Adam(\textrm{train}\underline{\hspace{0.2cm}}\textrm{loss})$\\
        
    }
}
        
        $\textrm{val}\underline{\hspace{0.2cm}}\textrm{loss}=\frac{1}{N\cdot\rho}\sum_{k=1}^{N\cdot\rho}{(f^L(\by_{t^0}^{[k]},\cdots,\by_{t^T}^{[k]};N_{\textrm{in}},N_{\textrm{out}},\textbf{W}) -\bR_{\textrm{true}}^{[k]})}^2$ 
        
        \If{$n_{\textrm{epoch}} == 0$}
        {
            $\textrm{min}\underline{\hspace{0.2cm}}\textrm{val}\underline{\hspace{0.2cm}}\textrm{loss}=\textrm{val}\underline{\hspace{0.2cm}}\textrm{loss}$\\
            
            $n\underline{\hspace{0.2cm}}\textrm{patience}\underline{\hspace{0.2cm}}\textrm{epoch}=0$\\
        }
        
        \If{$\textrm{val}\underline{\hspace{0.2cm}}\textrm{loss}<\textrm{min}\underline{\hspace{0.2cm}}\textrm{val}\underline{\hspace{0.2cm}}\textrm{loss}$}
        {
        $\textrm{min}\underline{\hspace{0.2cm}}\textrm{val}\underline{\hspace{0.2cm}}\textrm{loss}=\textrm{val}\underline{\hspace{0.2cm}}\textrm{loss}$\\
        }
        \Else
        {
            $n\underline{\hspace{0.2cm}}\textrm{patience}\underline{\hspace{0.2cm}}\textrm{epoch}+=1$\\
            \If{$n\underline{\hspace{0.2cm}}\textrm{patience}\underline{\hspace{0.2cm}}\textrm{epoch}==N_{\textrm{patience}\underline{\hspace{0.2cm}}\textrm{epoch}}$}
            {
                Terminate Algorithm
            }
        }

}

\caption{LSTM training and validating process for observation error covariance matrix}
\label{ALgo:LstmForR}
\end{algorithm}

In Algorithm \ref{ALgo:LstmForR}, 
it is suggested that LSTM training process is consisted of training and validation processes: the training process is comprised of the forward prediction and the backward neural network weight parameters updating processes; and validation process is used to predict desirable outputs or objectives and then calculate validation loss between predicted output and prior true output values. 
 $N_{\textrm{epoch}}$ indicates the number of times that the entire example data set is passed forward and backward through the LSTM during the training process. \ming{Early stopping, which terminates the training process when the validation loss reaches the minimum and is always the minimum value after $N_{\textrm{patience}\underline{\hspace{0.2cm}}\textrm{epoch}}$ epochs, is applied to reduce the LSTM training time.} \\
 
It is important to note that the offline data generation and LSTM training processes need to be carried out individually for different DA applications.

\begin{figure}
  \centering
  \includegraphics[width=7.0 in]{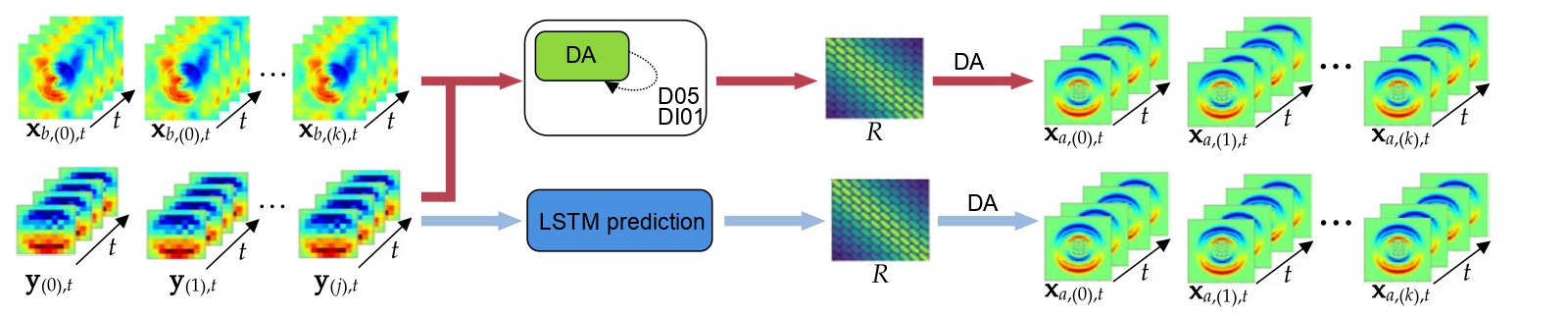}
    \caption{Online LSTM prediction of the observation matrix compared to traditional covariance tuning approaches, followed by online data assimilation}
    \label{fig:data_generation_prediction}
\end{figure}

\section{Lorenz twin experiment}
\label{sec:lorenz}

\subsection{Twin experiment principle}
\label{sec:twin experiment}
In order to overcome the drawback that, in a realistic experiment, $\bx_\textrm{true}$ is usually unknown and $\by$ is often mixed with noises, twin experiment, in which a prototypical test case is selected to simulate real situations, is applied so as to provide $\bx_\textrm{true}$ for comparison. In this experiment, a mapping is applied to some sampling true trajectory $\bx_{\textrm{true},t^k}$ at some points in space and time and arbitrary random noises are added to obtain simulated raw measurements $\by_{t^k}$. DA is then implemented starting from initial background state $\bx_{b,t^0}$ representing the prior information that could be obtained about corresponding state $\bx_{\textrm{true},t^0}$, along with initial raw measurement $\by_{t^0}$. The output state is then compared against $\bx_\textrm{true}$, verifying the distance of these two states and minimizing it to evaluate and improve the performance of DA. In this section, we use a twin experiment to evaluate the performance of applying DA to a simple Lorenz system in which raw measurement error covariance is estimated/adjusted using respectively DI01, D05 and LSTM.    

\subsection{Experiment set up}
\label{sec:lorenz_exp_set_up}
The Lorenz system, first studied by Edward Lorenz, is a system of ordinary differential equations. For certain parameter values and initial conditions, the Lorenz system is notable for having chaotic solutions, in particular the Lorenz attractor, toward which a system tends to evolve. 
The Lorenz 96 system\cite{Lorenz96} has been widely used as a prototypical test case to compare the performance of DA algorithms\cite{Suraj2020,Ouala2020,Arcucci2021}.
Here we build a twin experiment framework with a simple three dimensional Lorenz system in which the state vector is denoted as $\bx = [x_{(0)}, x_{(1)}, x_{(2)}] $. The studied Lorenz system can be characterized as:

\begin{align}
\label{eq:lorenz_system}
    \frac{\partial x_{(0)}}{\partial t}&=\sigma(x_{(1)}-x_{(0)})  \\
    \frac{\partial x_{(1)}}{\partial t}&=\alpha x_{(0)}-x_{(1)}-x_{(0)}x_{(2)}  \notag\\ 
    \frac{\partial x_{(2)}}{\partial t}&=x_{(0)}x_{(1)}-\beta x_{(2)}. \notag 
\end{align}
where $\partial t=0.001s$, $\sigma=10$, $\alpha=28$ and $\beta=2.667$.

The initial values of the true state $\textbf{x}_{\textrm{true},t^0}$ are set to be $[0, 1, 1.05]$ while the initial background state $\textbf{x}_{b,t^0}$ is generated by combining $\textbf{x}_{\textrm{true},t^0}$ with a centered Gaussian noise $\epsilon_{b,t^0}$: 
\begin{align}
    \textbf{x}_{b,t^0} = \textbf{x}_{\textrm{true},t^0} + \epsilon_{b,t^0} \quad \textrm{where} \quad \epsilon_{b,t^0} \sim \mathcal{N} \Big( 0, 0.05 \times \bI_3 \Big).
\end{align} 
 Then both of true states $\textbf{x}_{\textrm{true}}=\{\textbf{x}_{\textrm{true},t^0},\cdots,\textbf{x}_{\textrm{true},t^T}\}$ and background states $\textbf{x}_{b,t^k}=\{\textbf{x}_{b,t^0},\cdots,\textbf{x}_{b,t^T}\}$ of the Lorenz system evolve by conforming respectively to the Lorenz equation in Eq.~\ref{eq:lorenz_system} until t=1s with total $T=1000$ time steps.  

Subsequently, observations $\textbf{y}=\{\by_{t^0},\cdots,\by_{t^T}\}$ can be acquired by mapping $\textbf{x}_\textrm{true}$ through a linear observation operator
\begin{align}
    \bH = \begin{bmatrix}
        1 & 1 & 0 \\ 
        2 & 0 & 1\\
        0 & 0 & 3 
    \end{bmatrix},
\end{align}
and adding noises respecting multinormal distribution $\mathcal{N}(0,\bR)$ where $\bR$ is randomly generated following the process described in Sect.~\ref{sec:DA with LSTM}. 

EnDA is then applied in this twin experiment to update the background ensemble using available observations. More precisely, every 10 time steps, EnDA is applied on $\textbf{x}_{b,t^k}$ along with $\textbf{y}_{t^k}$ to obtain the analysis states $\textbf{x}_{a,t^k}$. To simulate future background states before the next assimilation step, we add some artificial noises $\epsilon_{b,t^{k+1}}$ to $\textbf{x}_{b,t^{k+1}}$ respecting multivariate distributions, i.e.,
\begin{align}
    \textbf{x}_{b,t^{k+1}}& = \mathcal{M}_{t^k \rightarrow t^{k+1} }(\textbf{x}_{a,t^k} + \epsilon_{b,t^{k}}) ,\notag \\
    \textbf{x}_{b,t^{(k+\gamma)}}& = \mathcal{M}_{t^{(k+\gamma-1)} \rightarrow t^{(k+\gamma)} }(\textbf{x}_{b,t^{(k+\gamma-1)}}) \quad \textrm{for} \quad \gamma \in \{ 2,...,10\}, \label{eq:xb_propa}
\end{align}
where 
\begin{align}
 \epsilon_{b,t^{k+1}} \sim  \mathcal{N}(0, \bQ) \quad \textrm{and} \quad \bQ = \begin{bmatrix}
        1 & 0.2 & 0 \\ 
        0.2 & 1 & 0.2\\
        0 & 0.2 & 1 
    \end{bmatrix}. \notag
\end{align}
The model error covariance matrix $\bQ$ is supposed to be time-invariant for all generated trajectories.

\subsection{DA with LSTM-based covariance estimation}
\label{sec:DA with LSTM}
The Sect.~\ref{sec:lorenz_exp_set_up} exhibits the process of generating artificial training data for the LSTM model. More details about the $\bR$ matrix generation, the outputs of the LSTM model within the training data are revealed in this section. $\bR$ are parameterized by four real coefficients $r_0$, $r_1$, $r_2$ and $v_\textbf{R}$ which determine respectively the three correlation coefficients and the error amplitude, i.e.,
\begin{align}
    \textbf{R}= v_\textbf{R} \times \begin{bmatrix}
        1 & r_0 & r_1 \\ 
        r_0 & 1 & r_2\\
        r_1 & r_2 & 1.
    \end{bmatrix}
\end{align}
In this study, $v_\textbf{R}$ is generated uniformly between 0 and 100, i.e., $v_\textbf{R} \sim \mathcal{U}(0,100)$ with $\mathcal{U}$ denoting the uniform probability distribution. The correlation coefficients $\{r_0,r_1,r_2\} \in (-1,1)^3$ are obtained via randomly generated SPD matrices \footnote{https://scikit-learn.org/stable/modules/generated/sklearn.datasets.make\underline{\hspace{0.15cm}}spd\underline{\hspace{0.15cm}}matrix.html}.

The LSTM network is thus trained to learn the $\bR$ matrix by trying to build a function mapping time series observation data $\by_{t^k}$ to $r_0$, $r_1$, $r_2$ and $v_\textbf{R}$. 
The specific structure of the LSTM model which consists of a LSTM input layer, a hidden layers with 200 neurons, and an output layer comprising four neurons applied to obtain $\textbf{R}$, is shown in Table~\ref{table: lstm_structure_lorenz}. 
In this Lorenz twin experiment, two LSTM networks, respectively named as  LSTM1000 and LSTM200 are designed with different input size of times series data. LSTM1000 is trained on a total of 1000 time steps for predicting the $\bR$ matrix while LSTM200 makes use of only the first 200 time steps to simulate a realistic application where the time-invariant $\bR$ matrix is estimated using historical data for improving future DA performance. The evaluation of both LSTM1000 and LSTM200, in terms of DA accuracy, is made using the full test dataset with 1000 times steps.
By leveraging the LSTM model along with available observations, we can still perform DA algorithms even though $\bR$ is not explicitly given. The results are then compared with the ones obtained using the exact $\bR$ matrix.

\begin{table}
\centering
\begin{tabular}{SSS} \toprule
    {\textbf{Layer (type)}} 
    
    & {\textbf{Intput Shape}} 
    
    & {\textbf{Output Shape}} \\ \midrule
    {lstm \textbf{(}LSTM\textbf{)}}  
    
    & \textbf{(}example\underline{\hspace{0.2cm}}size, time\underline{\hspace{0.2cm}}steps,3\textbf{)} 
    
    & \textbf{(}example\underline{\hspace{0.2cm}}size,200\textbf{)}\\
    {dense\textbf{(}Dense\textbf{)}}  
    
    & \textbf{(}example\underline{\hspace{0.2cm}}size,200\textbf{)}  
    
    & \textbf{(}example\underline{\hspace{0.2cm}}size,4\textbf{)} \\
    \bottomrule
\end{tabular}
\caption{Lstm specific structure for $\textbf{R}$ prediction in lorenz experiment}
\label{table: lstm_structure_lorenz}
\end{table}

\subsection{Results}
\ming{To evaluate the LSTM performance of predicting $\textbf{R}$, we first compare the LSTM predicted $\textbf{R}$ and the predefined true $\textbf{R}$ matrix in the test set and analyze the impact of predicted $\bR$} on DA accuracy. \sibo{Since the LSTM prediction of $\bR$ matrix is non-parametric in these twin experiments, we compare each element of the predicted matrix (i.e.,$r_0$, $r_1$, $r_2$ and $v_\textbf{R}$) against the ground truth.
As for the DA accuracy, we calculate the difference between $\bx_{b,t}=\{x_{b,(0),t},x_{b,(1),t},x_{b,(2),t}\}$} (obtained via Eq.~\ref{eq:xb_propa}) refined in DA using $\bR$ estimated via DI01 ($q=2$), D05 ($q = 3$) or LSTM and the error-free true states $\bx_{\textrm{true},t}=\{x_{\textrm{true},(0),t},x_{\textrm{true},(1),t},x_{\textrm{true},(2),t}\}$. Both D05 and DI01 initialize with a random $\bR$ matrix, generated through $g^{\bR}(.)$ with the same range of parameters $\Phi_\textbf{R}$ as for the LSTM training.

\begin{figure}

\centering

\subfloat[LSTM1000: $r_0$]{\label{fig:1000_1000_r0}\includegraphics[width = 1.6in]{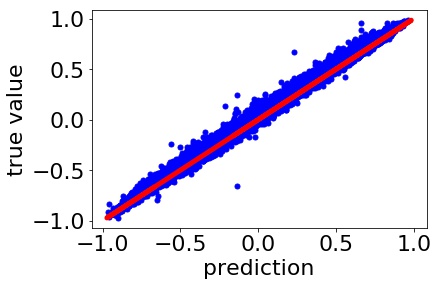}} 
\subfloat[LSTM1000: $r_1$]{\label{fig:1000_1000_r1}\includegraphics[width = 1.6in]{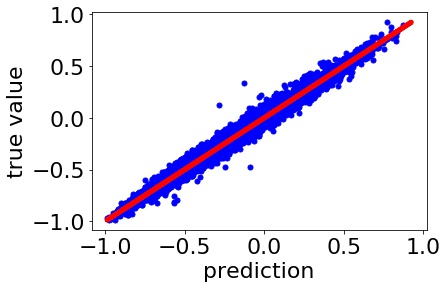}}
\subfloat[LSTM1000: $r_2$]{\label{fig:1000_1000_r2}\includegraphics[width = 1.6in]{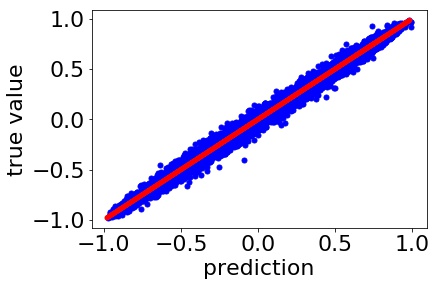}}
\subfloat[LSTM1000: $v_\textbf{R}$]{\label{fig:1000_1000_r3}\includegraphics[width = 1.6in]{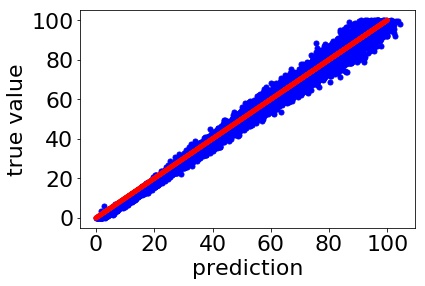}} \\

\subfloat[LSTM200: $r_0$]{\label{fig:200_200_r0}\includegraphics[width = 1.6in]{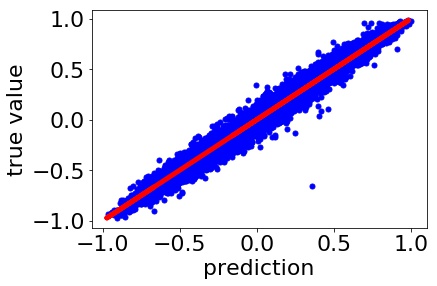}} 
\subfloat[LSTM200: $r_1$]{\label{fig:200_200_r1}\includegraphics[width = 1.6in]{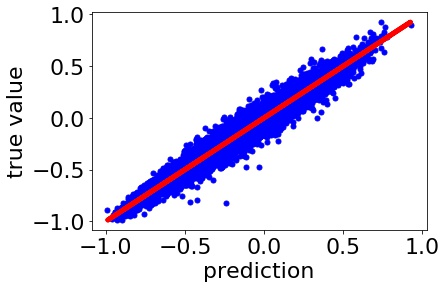}}
\subfloat[LSTM200: $r_2$]{\label{fig:200_200_r2}\includegraphics[width = 1.6in]{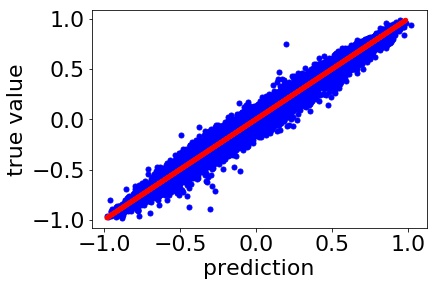}}
\subfloat[LSTM200: $v_\textbf{R}$]{\label{fig:200_200_r3}\includegraphics[width = 1.6in]{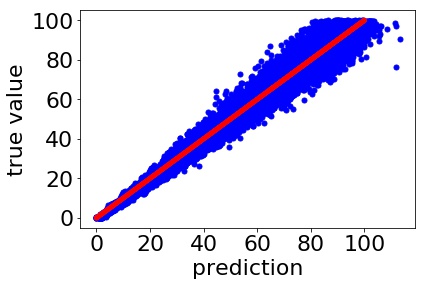}} \\

\subfloat[D05: $r_0$]{\label{fig:d05_r0}\includegraphics[width = 1.6in]{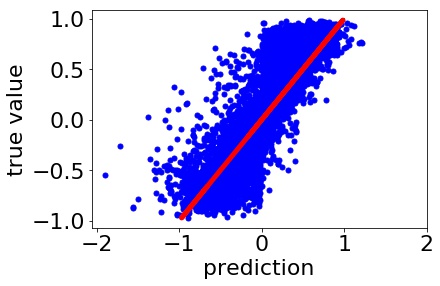}} 
\subfloat[D05: $r_1$]{\label{fig:d05_r1}\includegraphics[width = 1.6in]{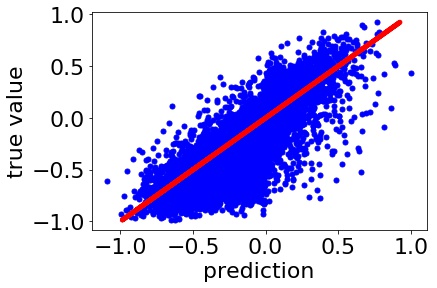}}
\subfloat[D05: $r_2$]{\label{fig:d05_r2}\includegraphics[width = 1.6in]{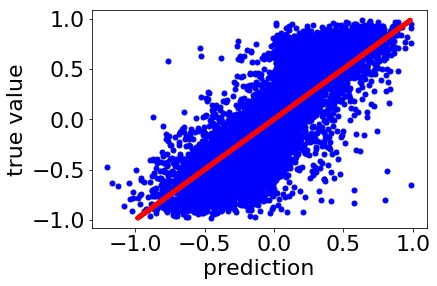}}
\subfloat[D05: $v_\textbf{R}$]{\label{fig:d05_r3}\includegraphics[width = 1.6in]{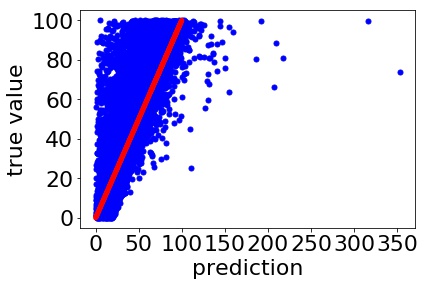}} 

   \caption{\sibo{Prediction results of the non-parametric error covariance and the true values in the test dataset of LSTM1000 (a-d), LSTM200 (e-h) and D05 (q = 2) (i-l) of $r_0,r_1,r_2$ and $v_\textbf{R}$ of the Lorenz system}}
   \label{fig:1000_1000_lstm_evaluation}
\end{figure}

Fig.~\ref{fig:1000_1000_lstm_evaluation} shows the elements of the estimated $\bR$ matrix (i.e.,$r_0$, $r_1$, $r_2$ and $v_\textbf{R}$) obtained by LSTM1000 and LSTM200, both trained on 103486 Lorenz system observation samples, and predicts on a test dataset of 10000 samples. The result of the D05 approach, with a calibration of $r_0$, $r_1$, $r_2$ and $v_\textbf{R}$ \sibo{against the true value}, has also been displayed in Fig.~\ref{fig:1000_1000_lstm_evaluation} (i-l) \sibo{in the same test dataset for comparison purposes}. The blue circles are the LSTM/D05 prediction results, while the red line is the true value of the corresponding parameter. We observe that both LSTM1000 and LSTM200 prediction results fit very well to the predefined true value of each parameter, compared to the D05 tuning approach, especially for the error amplitude $v_\textbf{R}$ \sibo{which is of most importance in covariance specification.} \sibo{Based on these experimental results, we can conclude that the proposed LSTM approach is capable of predicting the observation matrix, in terms of both correlation coefficients and error amplitude}, when time series observation data $\by_{t^k}$ are given. 

\begin{figure}[tb!]
\centering
  \includegraphics[width=7. in]{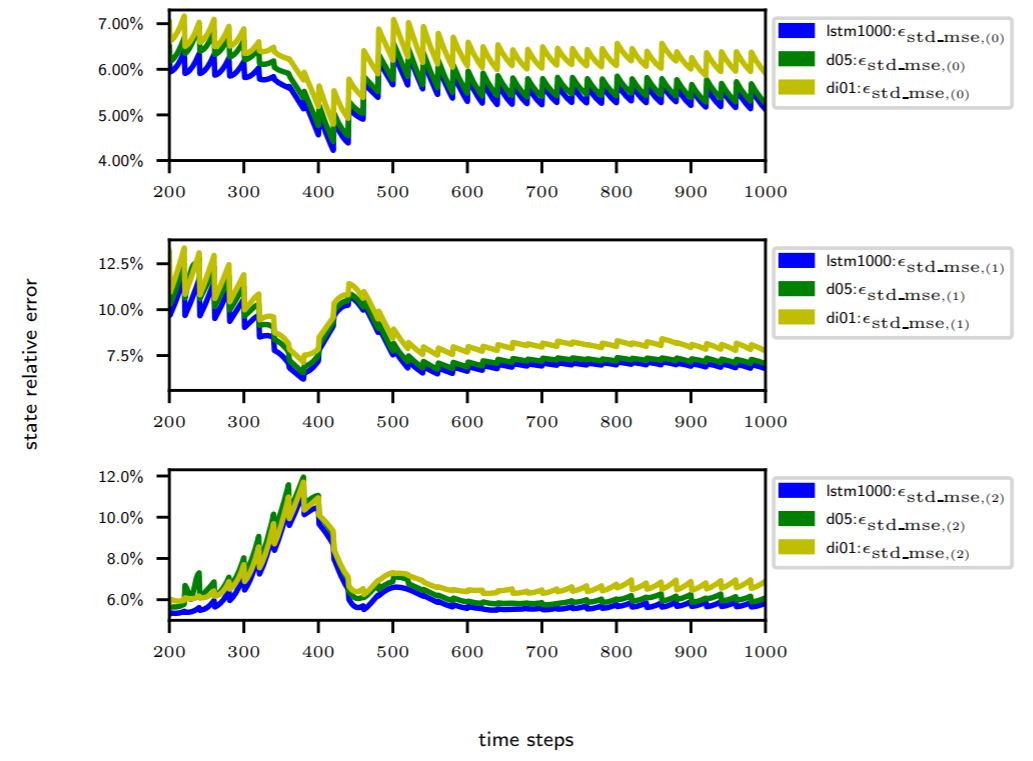}

\caption{DA performance evaluated in $\epsilon_{{\textrm{std}}\underline{\hspace{0.15cm}}{\textrm{mse}}}$ based on $\bR$ attained in algorithm LSTM, D05 and DI01}
\label{fig:1000_1000_da_performance}
\end{figure}

Fig.~\ref{fig:1000_1000_da_performance} and Table~\ref{table:epsion_mse} illustrate the averaged DA performance along with $\textbf{R}$ attained in different ways with 10000 observations in the test dataset. \ming{We remind that for each observation sample $\{ \by_{t^k}\}$, $t^{k}\in \{0,\cdots,t^T\}$}, 100 background trajectories are generated to perform EnDA. Among these algorithms, DI01 uses two iterations to correct the magnitudes of $\bB$ and $\bR$ by conforming to Eq.\ref{eq:B_R_iterated_update}. 
Each D05 iteration calculates the innovation quantities $(\textbf{y}-\mathcal{H}(\bx_a))$ and $(\textbf{y}-\mathcal{H}(\bx_b))$ every 10 time steps through a DA procedure, and then applies Eq.\ref{eq:it_D05} to attain the updated $\bR$ for the corresponding Lorenz system. 

Fig.~\ref{fig:1000_1000_da_performance} displays the evolution of the mean square error (MSE) $\epsilon_{{\textrm{std}}\underline{\hspace{0.15cm}}{\textrm{mse}},(i),t}$ between DA refined $\bx_{b,t}$ (following Eq.~\ref{eq:xb_propa}) and the true states $\bx_{\textrm{true},t}$, i.e.,
\begin{align}
    \epsilon_{{\textrm{std}}\underline{\hspace{0.15cm}}{\textrm{mse}},(i),t}=\sum_{j=1}^{N}\frac{\sqrt{\sum_{m=1}^{M}{(x_{b,(i),t}^{(m),[j]}-x_{(i),t}^{[j]})}^2}}{{\left\|\bx_{(i)}\right\|}_2}\Big/N
    \label{eq:epsilon_std_mse}
\end{align}
 where $i \in \{ 0,1,2\}$, $N=1988$ is the number of examples, and $M=100$ is the size of the ensemble DA.
Table~\ref{table:epsion_mse} shows the  averaged (against time) absolute error
\begin{align}
    \epsilon_{\textrm{mse},(i)}=\sum_{j=1}^{N}\frac{\sum_{t=t^0}^{t^{T}}\sqrt{\sum_{m=1}^{M}{(x_{b,(i),t}^{(m),[j]}-x_{(i),t}^{[j]})}^2}}{t^T} \Big/N
    \label{eq:mse}
\end{align}
to interpret the difference between assimilated and true states, where $t^T=1000$ is the total time steps that the Lorenz system has evolved.

It should be noted that the DI01 approach, which exclusively adjusts the $Tr(\bB)/Tr(\bR)$ ratio without modifying the correlation structure, is over-performed by more refined covariance tuning/specification methods such as D05 and LSTM \sibo{as shown in Fig.~\ref{fig:1000_1000_da_performance}}. \sibo{Furthermore, Table~\ref{table:epsion_mse} shows that $\textrm{lstm1000:}\epsilon_{{\textrm{std}}\underline{\hspace{0.15cm}}{\textrm{mse}}}$ is smaller than $\textrm{d05:}\epsilon_{{\textrm{std}}\underline{\hspace{0.15cm}}{\textrm{mse}}}$}, which is consistent with the results shown in Fig.~\ref{fig:1000_1000_lstm_evaluation}. Thus, we can conclude that LSTM1000 is sound at predicting $\bR$, contributing to a better DA performance, compared to DI01 and D05. 
Such a conclusion is further proved in Table~\ref{table:epsion_mse} in which $\epsilon_{\textrm{mse}}$ based on LSTM1000 \sibo{has values, for all three parameters displayed, close to those \ming{based on} true $\bR$}, which is predefined and used throughout the Lorenz system sample generations.

\begin{table}
\centering
\begin{tabular}{SSSSSSSSSS} \toprule
    {parameters} & {True} & {LSTM1000} & {LSTM200}   & {DI01} &  {D05}  \\ \midrule
    {$\epsilon_{\textrm{mse},(0)}$}  & 15.138 & 15.153 & 15.170 &  17.150 &  15.731 \\
    {$\epsilon_{\textrm{mse},(1)}$}  & 21.801  & 21.870 & 21.871 &   24.888 &  22.900 \\
    {$\epsilon_{\textrm{mse},(2)}$}  & 16.652  & 16.693 & 16.693 &  18.658 & 17.849\\
    \bottomrule
\end{tabular}

\caption{DA performance of the Lorenz system evaluated in $\epsilon_{\textrm{mse}}$ for the three state variables based on $\bR$ varied by the predefinition, the algorithm LSTM1000, LSTM200, DI01 and D05}
\label{table:epsion_mse}
\end{table}

What we have not exhibited in Fig.\ref{fig:1000_1000_da_performance} are the curves of $\textrm{lstm200:}\epsilon_{{\textrm{std}}\underline{\hspace{0.15cm}}{\textrm{mse}}}$ and $\textrm{true:}\epsilon_{{\textrm{std}}\underline{\hspace{0.15cm}}{\textrm{mse}}}$, representing the MSE based on LSTM200 predicted $\bR$ and manually predefined true $\bR$ respectively, as they have been almost totally overlapped with $\textrm{lstm1000:}\epsilon_{{\textrm{std}}\underline{\hspace{0.15cm}}{\textrm{mse}}}$. Such a fact permits both the same conclusion to be reached with LSTM1000, and that LSTM200 which makes use of the observation data of only the first 200 time steps, is sound at predicting $\bR$ to contribute to a good DA performance with future observations. This conclusion is supported in Table~\ref{table:epsion_mse} where $\epsilon_{\textrm{mse}}$ based on LSTM200 predicted $\bR$ has almost the same values as that based upon manually predefined true $\bR$. 

\section{Application to shallow water equations}
\label{sec:sw}
\subsection{Experiment set up}
For further evaluating the performance of error covariance estimation using LSTM when incorporated with predefined correlation kernels, we also set up a twin experiment framework with a simplified 2D shallow water dynamical model which is frequently used for testing data assimilation algorithms (
 e.g., \cite{Cioaca2014,cheng2019}). A cylinder of water is positioned in the middle of the study field with size $20mm \times 20mm$ and released at the initial time step $t^k=t^0s$ (i.e., with no initial speed), leading to a non-linear wave-propagation. The dynamics of the water level $h$ (in $mm$), as well as the horizontal and vertical velocity field (respectively denoted as $u$ and $v$ in $0.1m/s$), is given by the non-conservative shallow water equations,
 
   \begin{minipage}{.5\textwidth}
\begin{align}
\label{eq: shallow_water_equation}
    \frac{\partial u}{\partial t}&=-g\frac{\partial}{\partial x}(h)-bu  \\
    \frac{\partial v}{\partial t}&=-g\frac{\partial}{\partial y}(h)-bv  \notag\\ 
    \frac{\partial h}{\partial t}&=-\frac{\partial}{ \partial x}(uh)-\frac{\partial}{\partial y}(v h)  \notag \\
    u_{t^0} &= 0 \notag \\
    v_{t^0} &= 0 \notag
\end{align}

  \end{minipage}
  \hspace{10mm}
     \begin{minipage}[H]{.4\textwidth}
        \includegraphics[width=2. in]{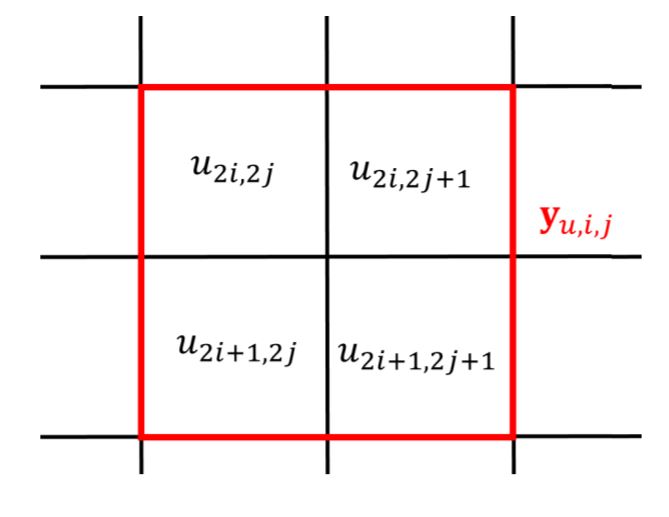}\\
        \captionof{figure}{State ($u,v$) - observation ($\mathbf{y}$) transformation}
        \label{fig:y_to_u}
  \end{minipage}\\
 
\begin{figure}[H]
  \centering
  \subfloat[$h(t^{800}=0.0s)$]{\includegraphics[width=2.2 in]{{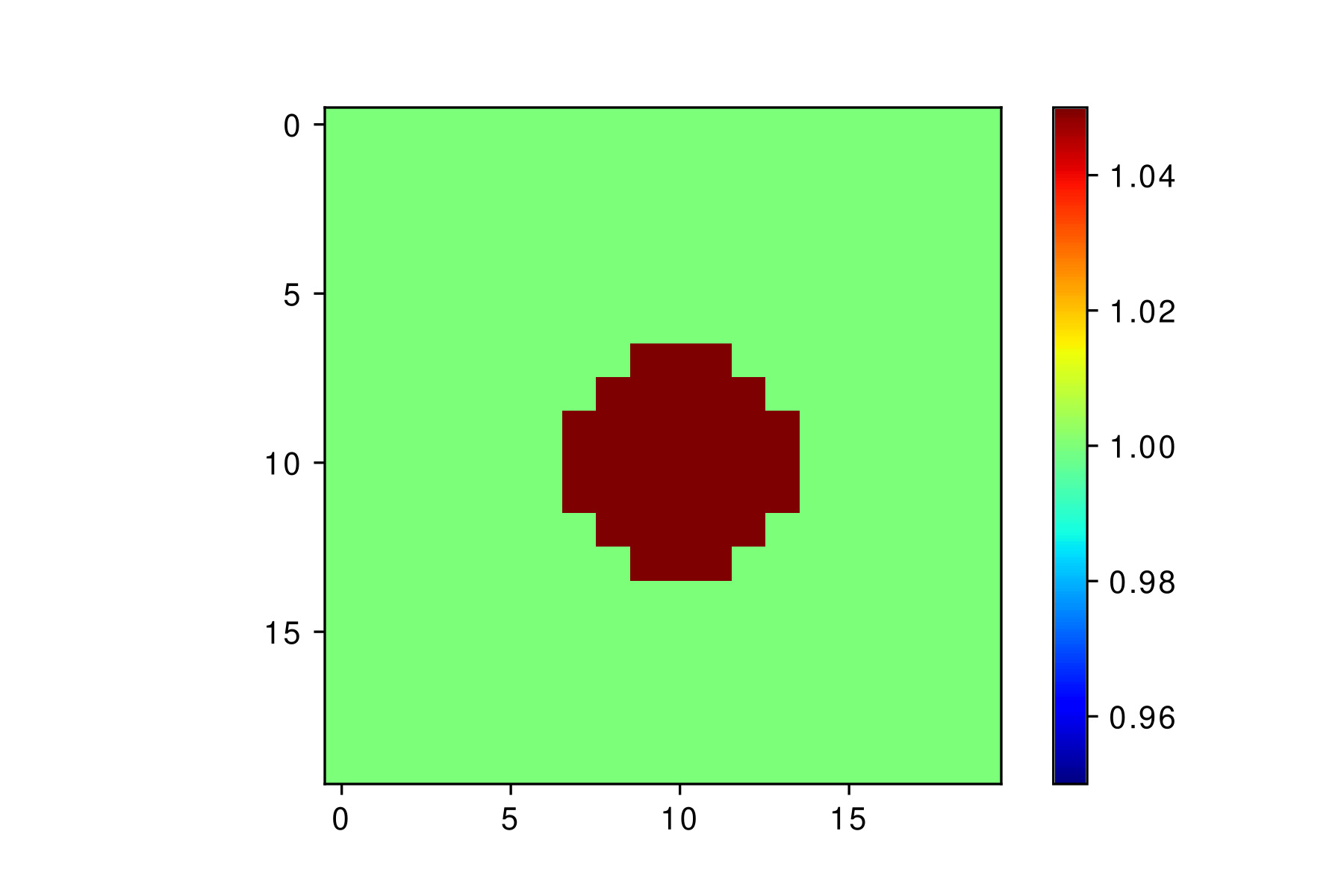}}}
  \subfloat[$u(t^{800}=0.8s)$]{\includegraphics[width=2.2 in]{{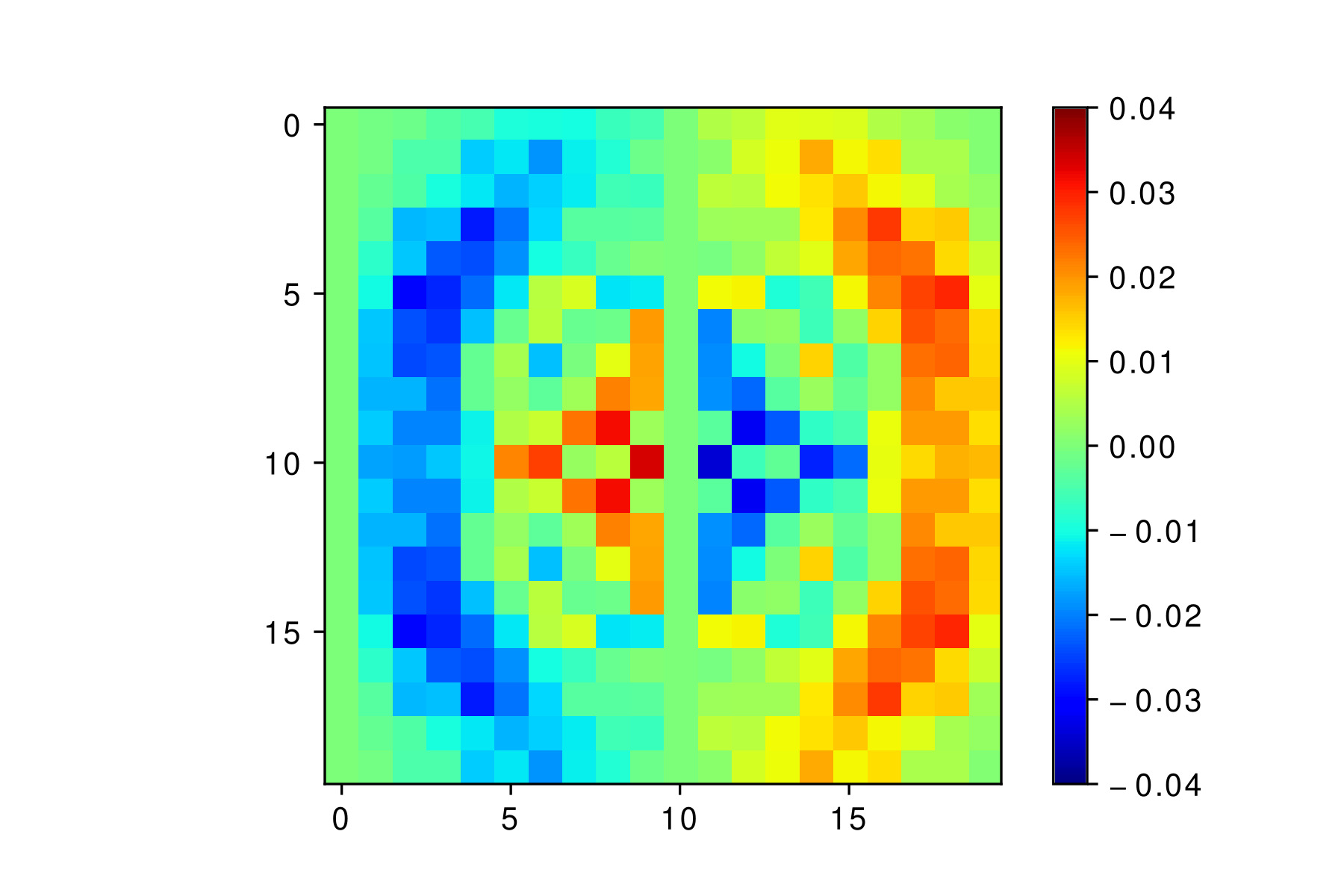}}}
 \subfloat[$\by_u(t^{800}=0.8s)$]{\includegraphics[width=2.2 in]{{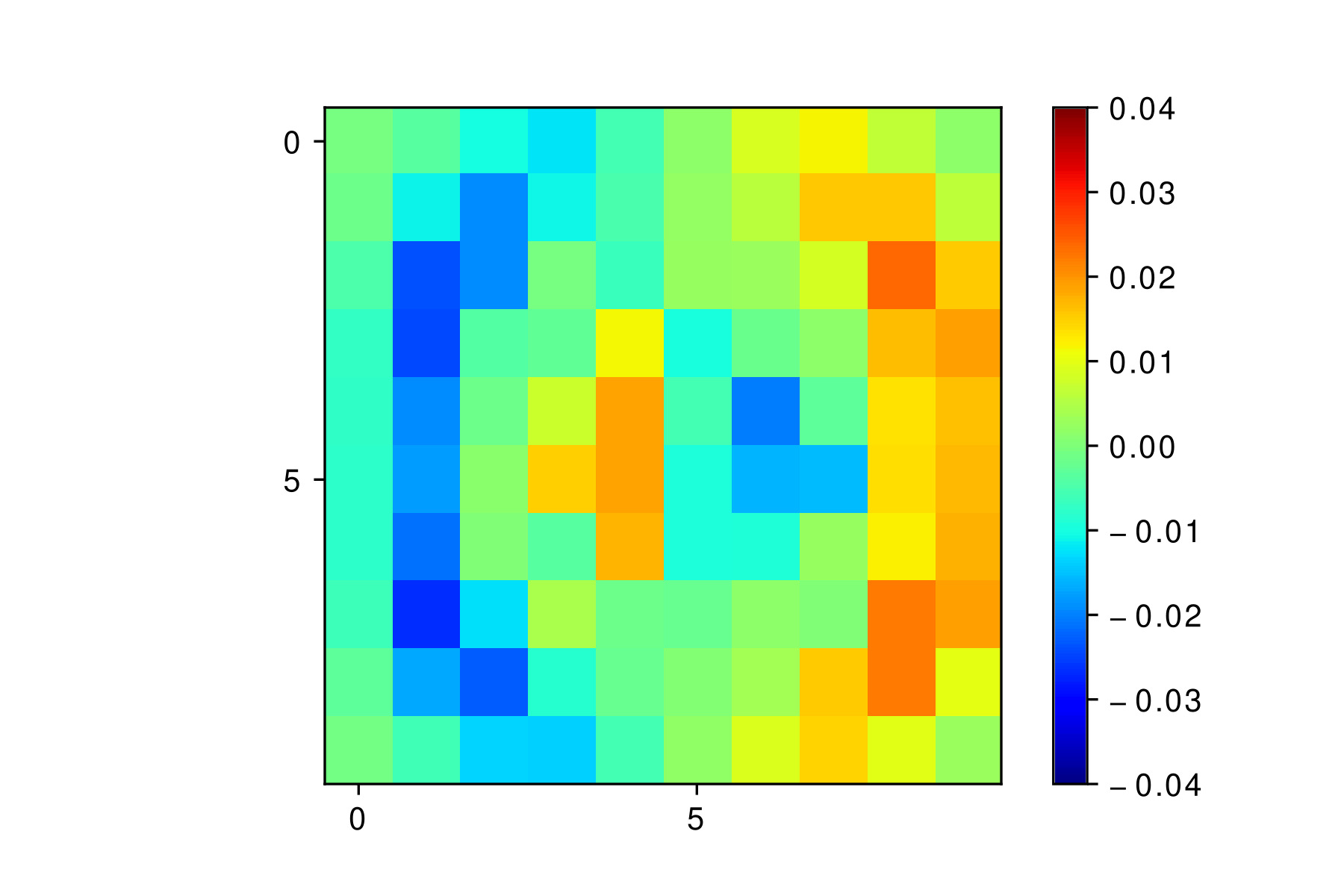}}}\\
    \subfloat[$h(t^{800}=0.08s)$]{\includegraphics[width=2.2 in]{{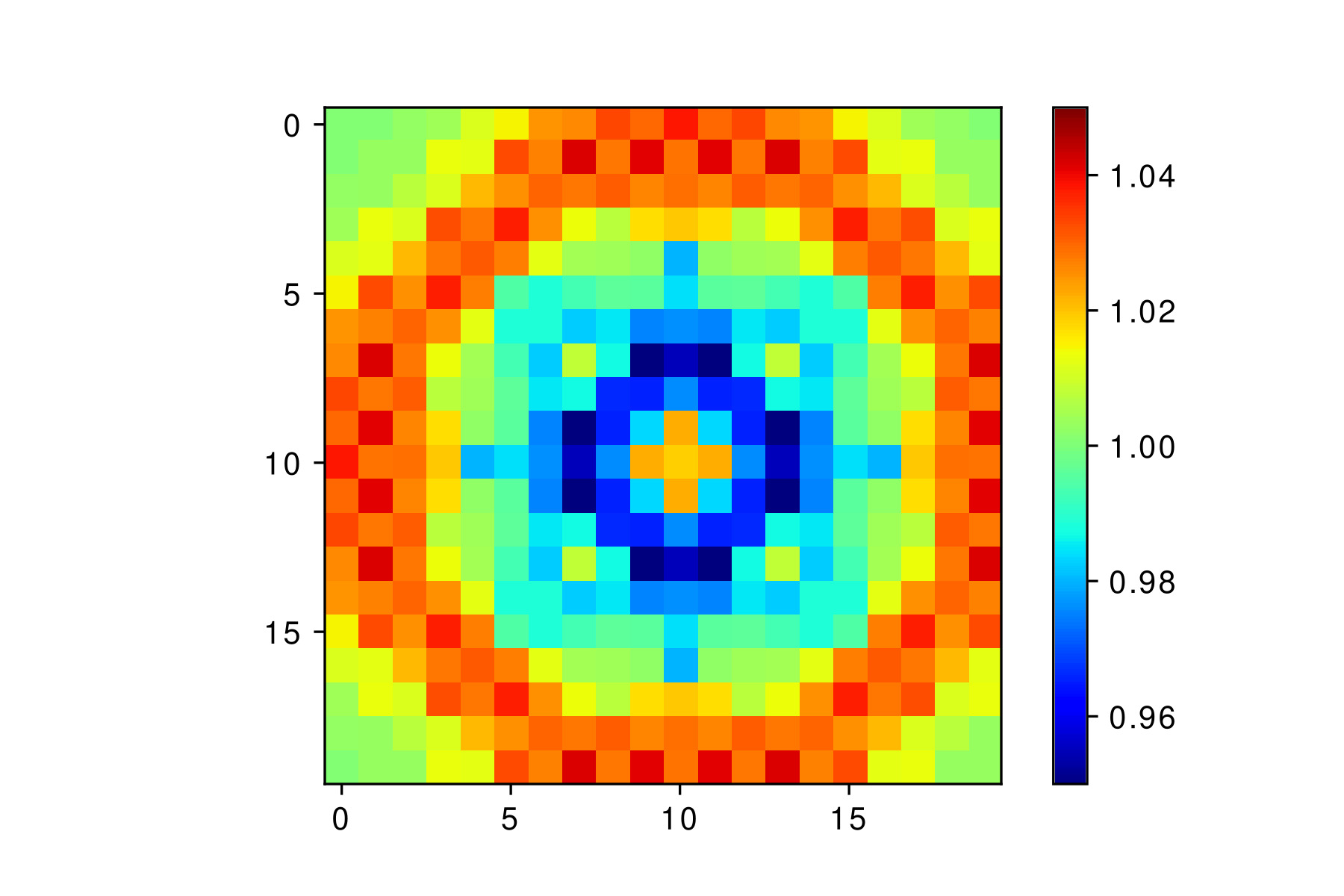}}}
    \subfloat[$v(t^{800}=0.08s)$]{\includegraphics[width=2.2 in]{{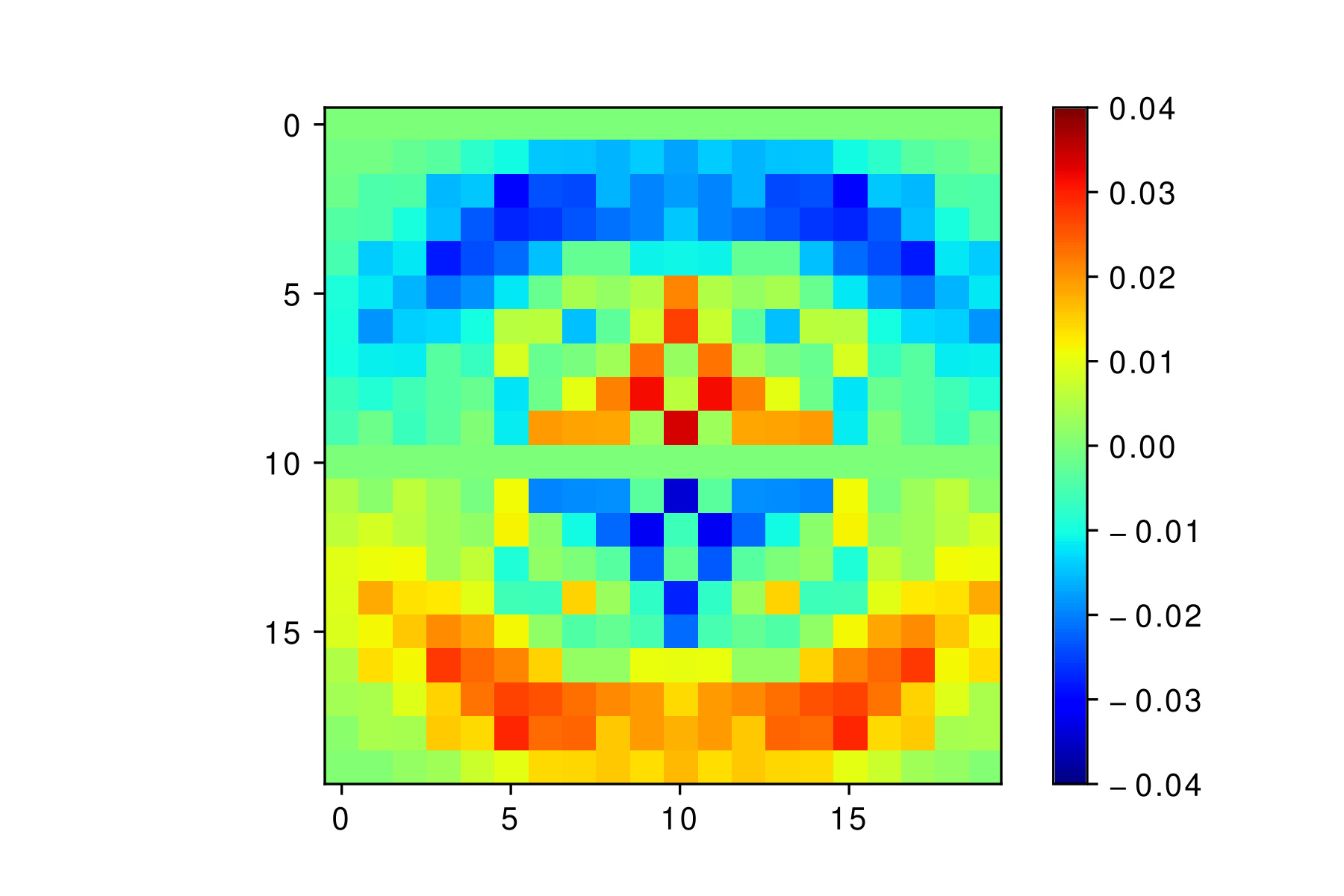}}}
    \subfloat[$\by_v(t^{800}=0.08s)$]{\includegraphics[width=2.2 in]{{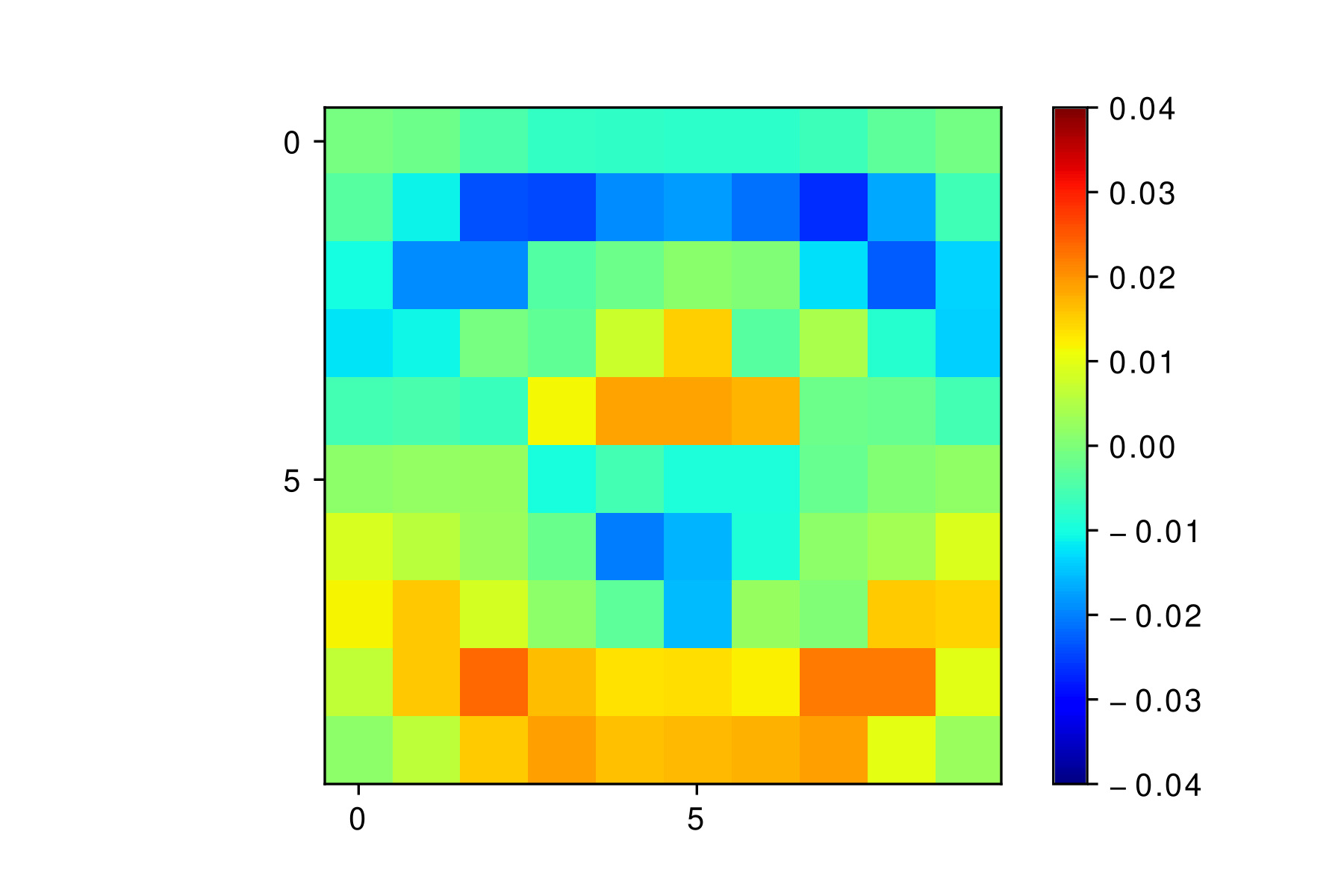}}}
    \caption{Evolution of the shallow water model of $h,u,v$ (true states) at different time steps (a,b,d,e) and the error-free model equivalent observations $\bH(\bx_{\textrm{true}})$ (c,f).}
  \label{fig:simulation}
\end{figure}

\vspace{4mm}
In Eq.\ref{eq: shallow_water_equation}, $b=0.1$ is the viscous drag coefficient, while $g$ is the constant earth gravity.
These equations are discretized in a $20 \times 20$ regular grid, solved by first-order finite difference method with a time discretization $\delta_t = 10^{-4}s$, from $t^0 = 0s$ to $t^{1000} = 0.1s$. This resolution is considered as the reference (i.e. the true state $\textbf{x}_\textrm{true}$) when performing DA algorithms. The state variables in this DA modeling are the combination of the velocity fields $\{u\}_{20 \times 20}$ and $\{v\}_{20 \times 20}$, leading to the state dimension of 800. The evolution of the reference state ($\textbf{x}_{\textrm{true},t^k}$), together with the error-free model equivalent observations (i.e., $\bH(\bx_\textrm{true})$),  is illustrated in Fig.~\ref{fig:simulation}. Spatially correlated prior observation errors are generated artificially and combined with the transformation operator to simulate real-time observations. More precisely, the observations are generated from the model equivalent $\mathbf{H}(\bx_\textrm{true})$ separately for the fields $u$ and $v$. $\textbf{H}$ is defined as a sparse matrix to imply the fact that measurements in real-world applications are sparser than true states due to the interference existing in the former situations as well as the limited performances of sensors. As shown in Fig.~\ref{fig:y_to_u}, the spatial observations at time \ming{$t^k$} is defined as the average of \ming{$u_{t^k}$} and \ming{$v_{t^k}$} in a $2 \times 2$ cells area with an observation error \ming{$\epsilon_{y_{t^k}}$}, 
\begin{align}
   \ming{\by_{u,i,j,t^k} = \frac{1}{4} (u_{\textrm{true},2i,2j,t^k} + u_{\textrm{true},2i+1,2j,t^k} + u_{\textrm{true},2i,2j+1,t^k} + u_{\textrm{true},2i+1,2j+1,t^k}) + \epsilon_{y_{u,i,j,t^k}},}
\end{align}
and identical for \ming{$\by_{v,i,j,t^k}$}. Therefore, the dimension of the observation vector \ming{$\by = [\by_{u,t^k}, \by_{v,t^k} ]$} is 200. In this experiment, we suppose that the observation error \ming{$\epsilon_{y_{u,i,j,t^k}}$} and \ming{$\epsilon_{y_{v,i,j,t^k}}$}, respectively of the velocity fields $u$ and $v$, follow the same Gaussian distribution $\mathcal{N}(0,\bR)$. Thus, the observation error covariance in this shallow water system can be fully characterized by a $100 \times 100$ $\bR$ matrix after the observations (originally in a 2D grid) being converted to a 1D vector.
Here we adopt a different parameterization of the $\textbf{R}$ matrix thanks to an isotropic correlation function $\psi_\mathbf{R}(.)$,
\begin{align}
\label{eq:R_definition}
    \bR=10^{-6} \cdot \sqrt{diag(\bD)}\cdot \psi_\mathbf{R} (r)\cdot\sqrt{diag(\bD)}  
\end{align}
where $\textbf{D} = \big[ D_0, ..., D_{99}\big]$, representing the error variances in the 2D ($10 \times 10$) velocity field. Each element of $\bD$ is generated individually following an uniform distribution, 
\begin{align}
\label{eq:d_definition}
    \bD_{\iota} \sim \mathcal{U}(1, 1000) \quad \textrm{for} \quad \iota \in \{ 0,...,99\},
\end{align} 
 which produces only positive elements to guarantee the positive definiteness of $\bR$. 

$\psi_\mathbf{R} (.)$ is the second-order auto-aggressive (also known as Balgovind) function,
\begin{align}
\label{eq:soar}
    \psi_\mathbf{R} (r) = \left ( 1+\frac{r}{L_\bR}\right ) \exp(-\frac{r}{L_\bR}),
\end{align}
where $L_\bR$ is the correlation scale length, fixed as $L_\bR = 10$ in this application. $r$ denotes the correlation scale length in the 2D space and is also generated uniformly with $r \sim \mathcal{U}(1,5)$. Being part of Matern kernels, the SOAR function is often used in DA for prior error covariance modeling \cite{cheng2019,Gong2020} thanks to its smoothness and good conditioning.
The simulation of \ming{$\textbf{x}_{b,t^k} = [{u}_{b,t^k}, {v}_{b,t^k}]$} via the same discretization of Eq.\ref{eq: shallow_water_equation} (except the initial conditions) is used as background states at time \ming{$t^k$} in the DA modeling. Similar to the Lorenz experiment(i.e., Eq.~\ref{eq:xb_propa}), \ming{$\{\textbf{x}_{b,t^k}\}$} is acquired by combining \ming{$\{\textbf{x}_{a,t^k}\}$} with randomly generated Gaussian errors, while \ming{$\{\textbf{x}_{a,t^k}\}$} is obtained every 100 time steps (i.e., $0.01s$) from ensemble DA with time series observation data \ming{$\{\textbf{y}_{t^k}\}$} and the estimated observation error covariance $\textbf{R}$.

\subsection{DA with LSTM estimated $\textbf{R}$}

As with the Lorenz experiment, simulated observations $\{\by_{t^k}\}$, generated in the same process with that in the Lorenz system, are used as input training data for the LSTM model,
while the $\textbf{D}$ vector and the correlation scale $r$ served as training output. 
The specific structure of this LSTM network is shown in Table~\ref{table: lstm_structure_shallow_water}. This model has the same structure as the one of the Lorenz system shown in Table~\ref{table: lstm_structure_lorenz}, except the input and ouput dimension.  
Besides, two types of LSTM are also proposed as what have been realized in Lorenz system: LSTM1000 employs the whole 1000 time steps observation data as LSTM training and prediction inputs while LSTM200 makes use of only the first 200 time steps of observation data as the LSTM inputs.   
After the LSTM is well trained on the training set of 173000 generated observation trajectories in this experiment, $\textbf{R}$ can be gained even only observation time series data $\{\by_{t^k}\}$,is acknowledged. 


\begin{table}
\centering
\begin{tabular}{SSS} \toprule
    {\textbf{Layer (type)}} 
    
    & {\textbf{Intput Shape}} 
    
    & {\textbf{Output Shape}} \\ \midrule
    {lstm \textbf{(}LSTM\textbf{)}}  
    
    & \textbf{(}example\underline{\hspace{0.2cm}}size,time\underline{\hspace{0.2cm}}steps,200\textbf{)} 
    
    & \textbf{(}example\underline{\hspace{0.2cm}}size,200\textbf{)}\\
    {dense\textbf{(}Dense\textbf{)}}  
    
    & \textbf{(}example\underline{\hspace{0.2cm}}size,200\textbf{)}  
    
    & \textbf{(}example\underline{\hspace{0.2cm}}size,101\textbf{)} \\
    \bottomrule
\end{tabular}
\caption{Lstm specific structure for $\textbf{R}$ prediction in shallow water experiment}
\label{table: lstm_structure_shallow_water}
\end{table}

Similar to the Lorenz system, EnDA is performed here with an ensemble of 100 state trajectories initialized from the same initial state $\bx_{t^0}$ for each observation series. EnDA takes place every 200 time steps with the $\bR$ matrix estimated through different methods.

\subsection{Results}
\begin{figure}[H]

\centering

\subfloat[$D_0: u/v(0,0)$]{\label{fig:1000_1000_sw_d0}\includegraphics[width = 1.6in]{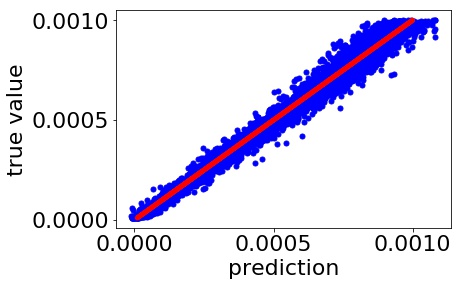}} 
\subfloat[$D_{24}: u/v(2,4)$]{\label{fig:1000_1000_sw_d24}\includegraphics[width = 1.6in]{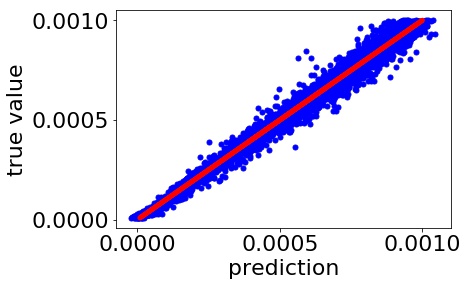}}
\subfloat[$D_{39}: u/v(3,9)$]{\label{fig:1000_1000_sw_d39}\includegraphics[width = 1.6in]{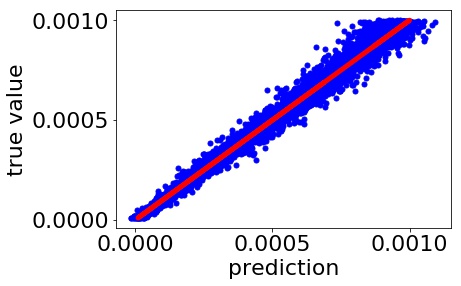}}
\subfloat[$D_{54}: u/v(5,4)$]{\label{fig:1000_1000_sw_d54}\includegraphics[width = 1.6in]{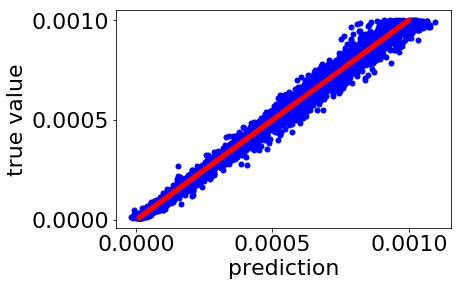}} \\

\subfloat[$D_{69}: u/v(6,9)$]{\label{fig:1000_1000_sw_d69}\includegraphics[width = 1.6in]{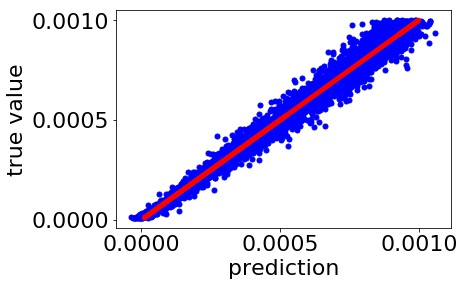}} 
\subfloat[$D_{84}: u/v(8,4)$]{\label{fig:1000_1000_sw_d84}\includegraphics[width = 1.6in]{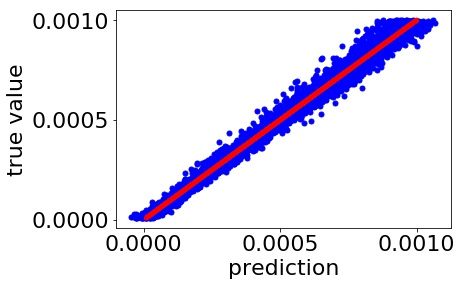}}
\subfloat[$D_{99}: u/v(9,9)$]{\label{fig:1000_1000_sw_d99}\includegraphics[width = 1.6in]{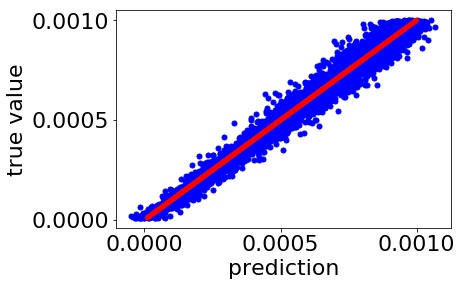}}
\hspace{3mm}
\subfloat[$r$]{\label{fig:1000_1000_sw_r}\includegraphics[width = 1.4in]{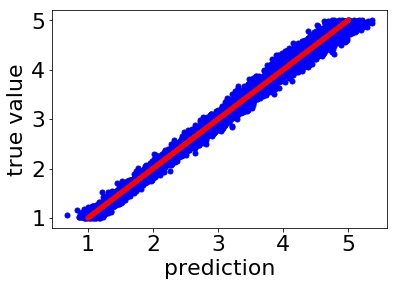}} 
   \caption{\sibo{Prediction results of the parametric error covariance and the true values in the test dataset of LSTM1000 for some elements of $\bD$ (representing marginal error variance in the 2D space)(a-g) and the correlation scale $r$ (h) of the shallow water system}}
   \label{fig:1000_1000_lstm_shallowW_evaluation}
\end{figure}

\begin{figure}[!ht]

\centering

\subfloat[$D_0: u/v(0,0)$]{\label{fig:200_200_sw_d0}\includegraphics[width = 1.6in]{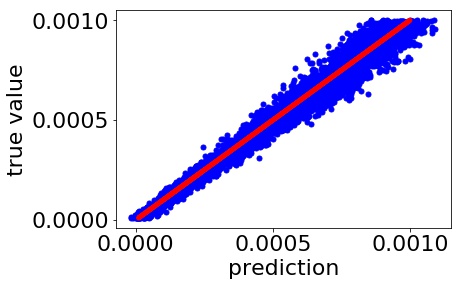}} 
\subfloat[$D_{24}: u/v(2,4)$]{\label{fig:200_200_sw_d24}\includegraphics[width = 1.6in]{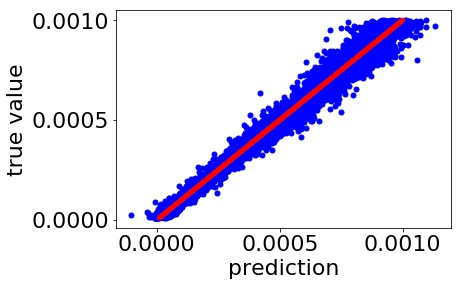}}
\subfloat[$D_{39}: u/v(3,9)$]{\label{fig:200_200_sw_d39}\includegraphics[width = 1.6in]{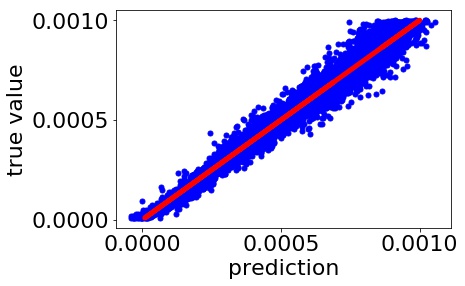}}
\subfloat[$D_{54}: u/v(5,4)$]{\label{fig:200_200_sw_d54}\includegraphics[width = 1.6in]{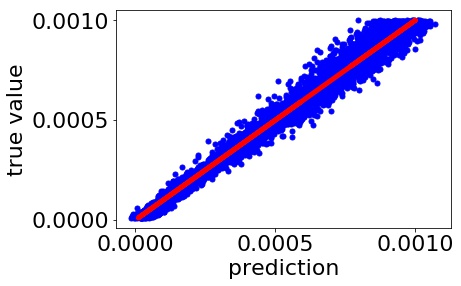}} \\

\subfloat[$D_{69}: u/v(6,9)$]{\label{fig:200_200_sw_d69}\includegraphics[width = 1.6in]{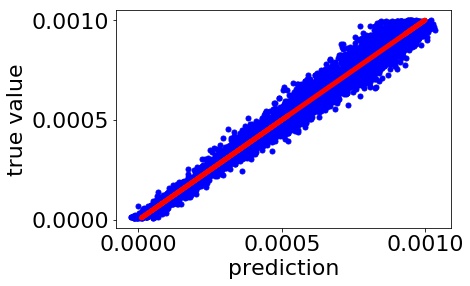}} 
\subfloat[$D_{84}: u/v(8,4)$]{\label{fig:200_200_sw_d84}\includegraphics[width = 1.6in]{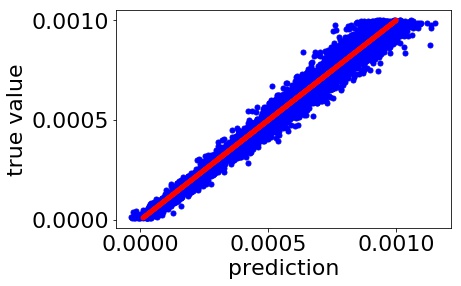}}
\subfloat[$D_{99}: u/v(9,9)$]{\label{fig:200_200_sw_d99}\includegraphics[width = 1.6in]{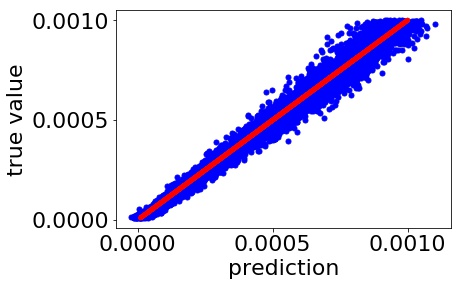}}
\hspace{3mm}
\subfloat[$r$]{\label{fig:200_200_sw_r}\includegraphics[width = 1.4in]{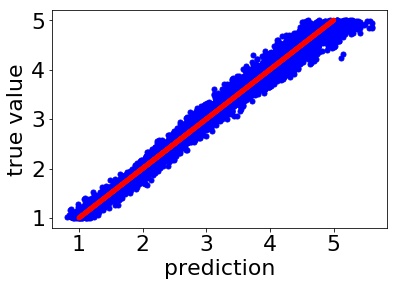}} 
   \caption{\sibo{Prediction results of the parametric error covariance and the true values in the test dataset of LSTM200 for some elements of $\bD$ (representing marginal error variance in the 2D space)(a-g) and the correlation scale $r$ (h) of the shallow water system}}
   \label{fig:200_200_lstm_shallowW_evaluation}
\end{figure}

\begin{figure}[!ht]
  \centering
  \subfloat[$\bR_\textrm{true}$: ex1]{\includegraphics[width=1.8 in]{{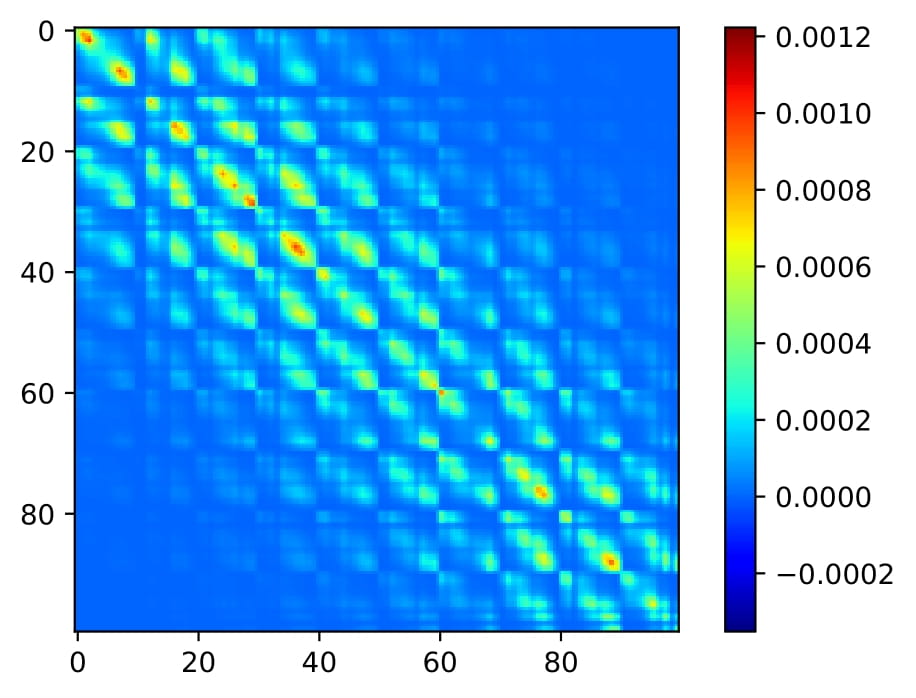}}}
    \subfloat[$\bR_\textrm{true}$: ex2]{\includegraphics[width=1.8 in]{{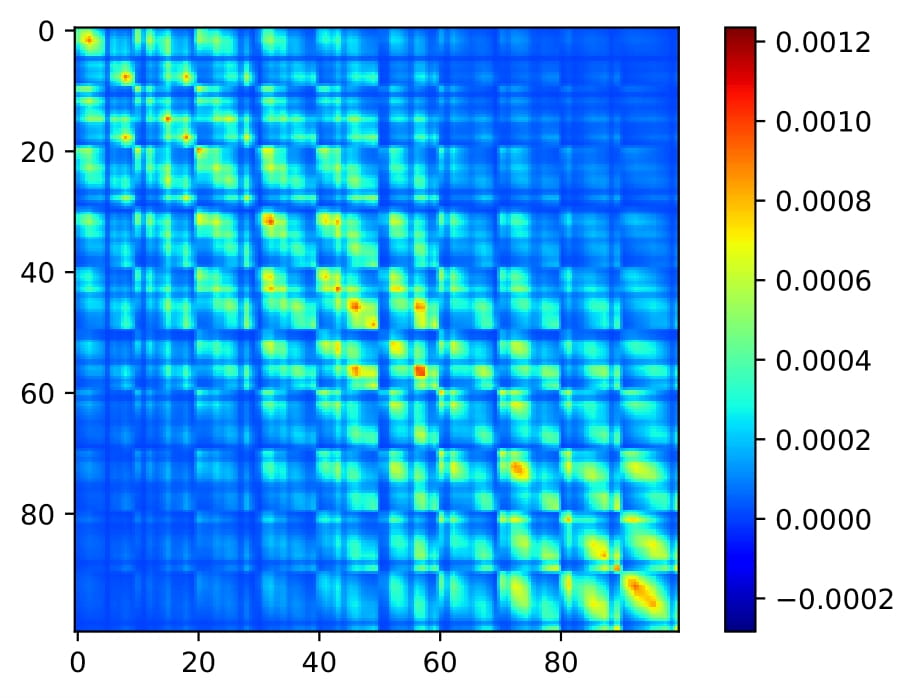}}}
      \subfloat[$\bR_\textrm{true}$: ex3]{\includegraphics[width=1.8 in]{{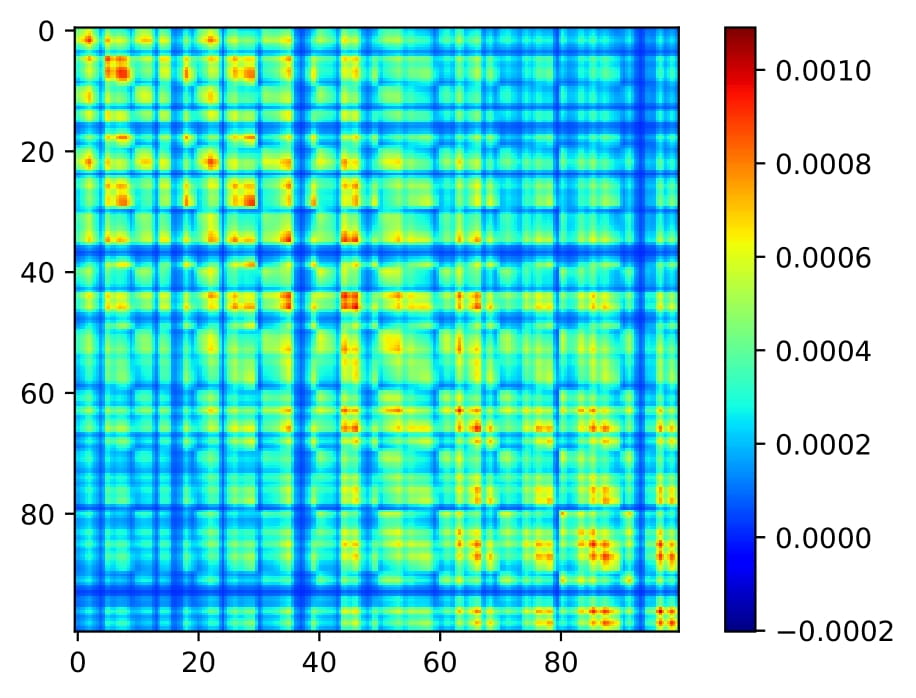}}}
        \subfloat[$\bR_\textrm{true}$: ex4]{\includegraphics[width=1.8 in]{{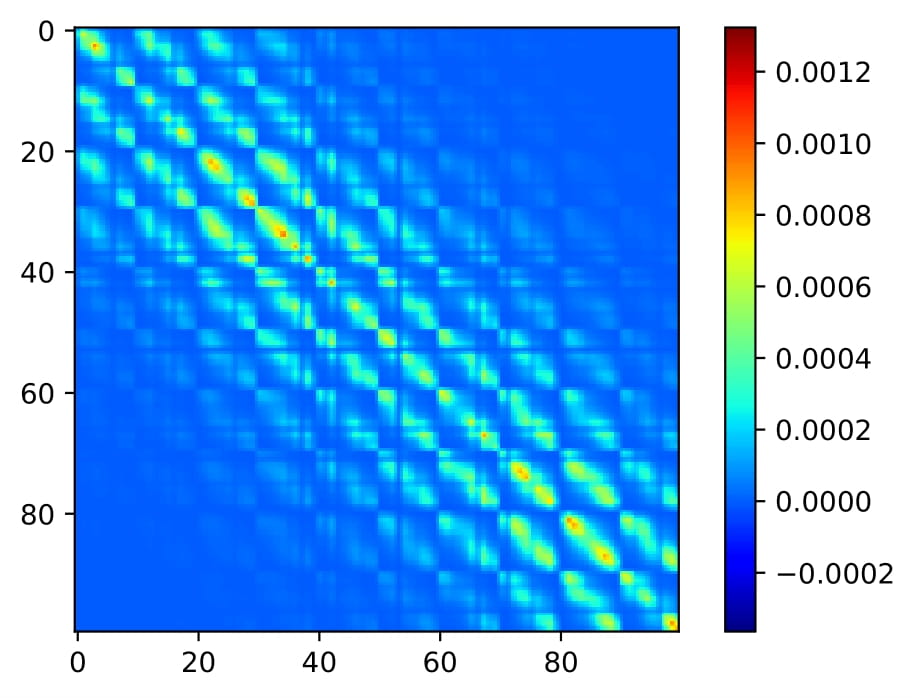}}}\\
  \subfloat[$\bR_\textrm{LSTM}$: ex1]{\includegraphics[width=1.8 in]{{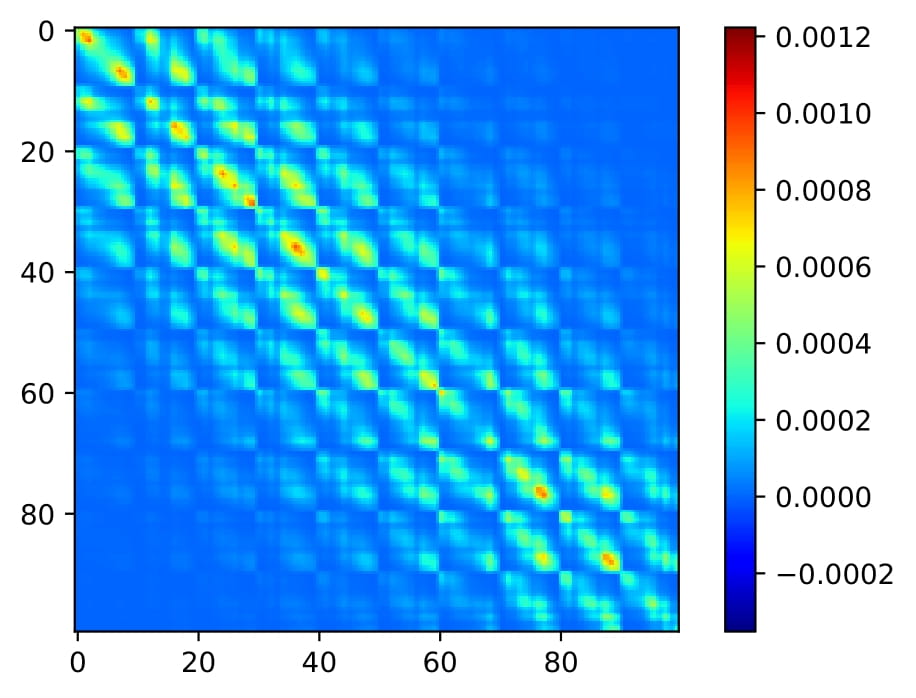}}}
    \subfloat[$\bR_\textrm{v}$: ex2]{\includegraphics[width=1.8 in]{{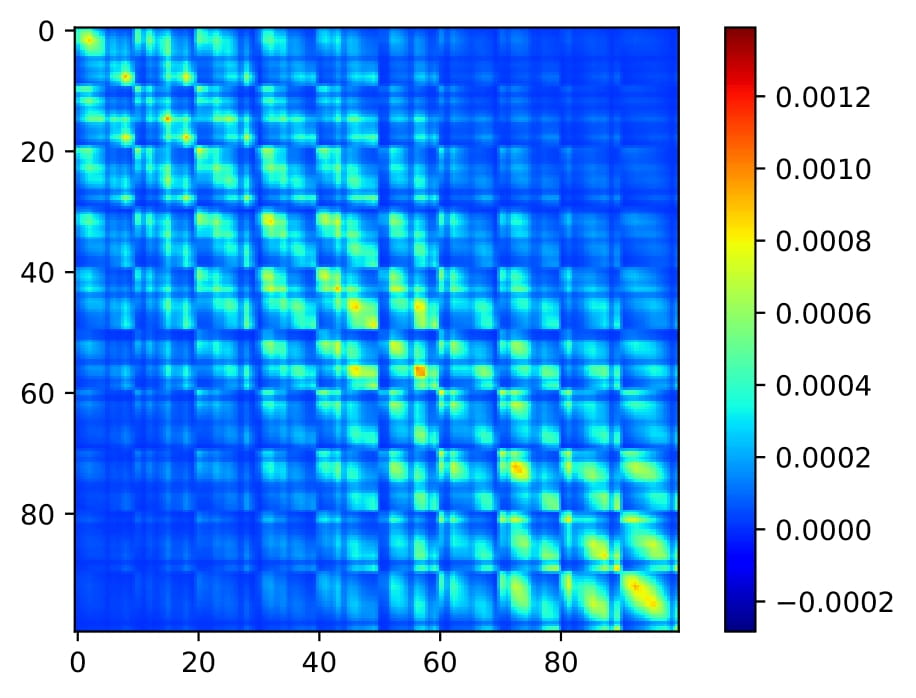}}}
      \subfloat[$\bR_\textrm{LSTM}$: ex3]{\includegraphics[width=1.8 in]{{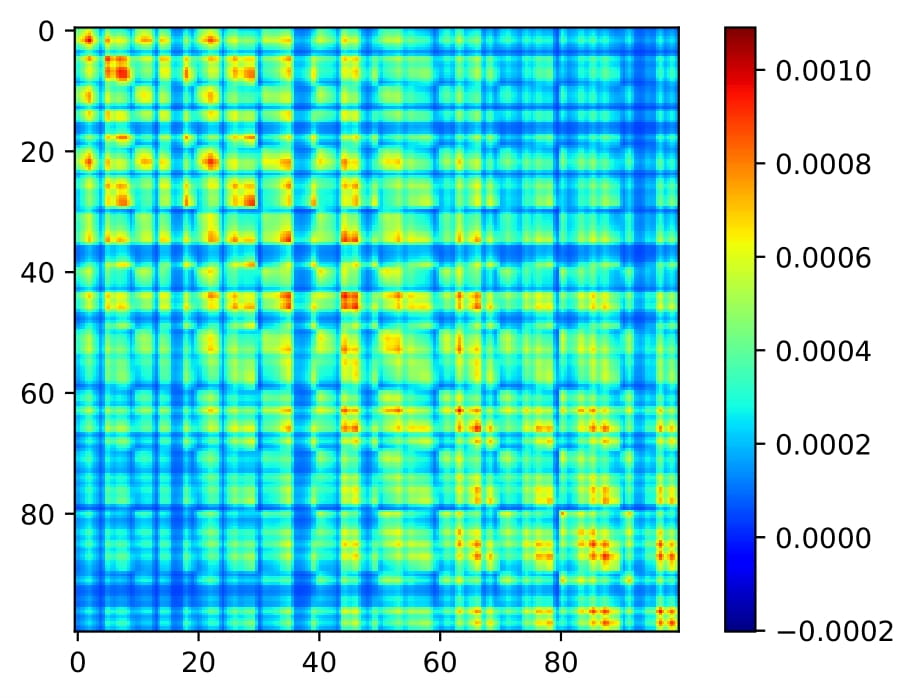}}}
        \subfloat[$\bR_\textrm{LSTM}$: ex4]{\includegraphics[width=1.8 in]{{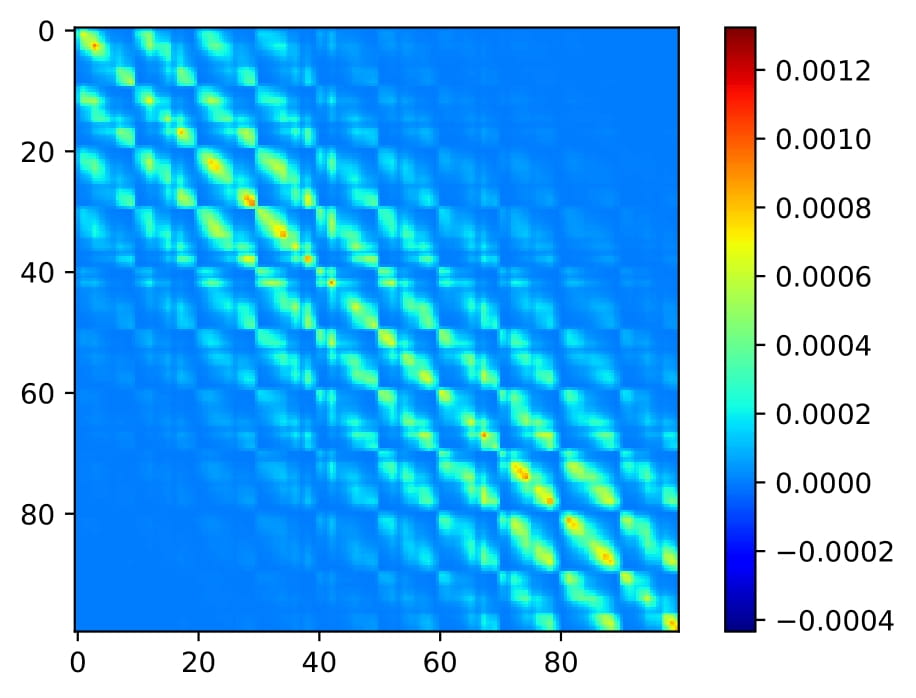}}}\\
  \subfloat[$\bR_\textrm{D05}$: ex1]{\includegraphics[width=1.8 in]{{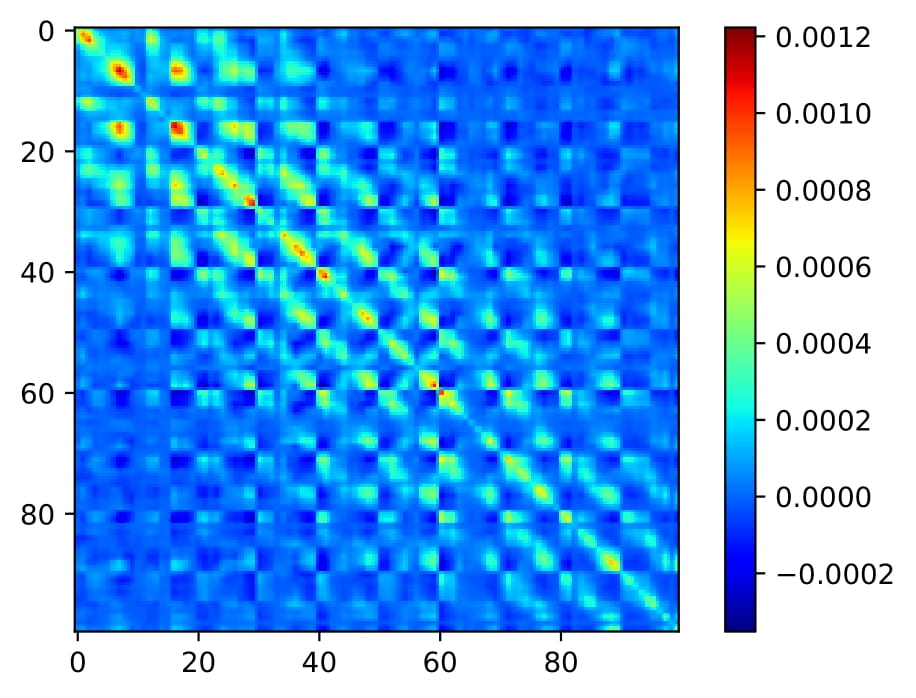}}}
    \subfloat[$\bR_\textrm{D05}$: ex2]{\includegraphics[width=1.8 in]{{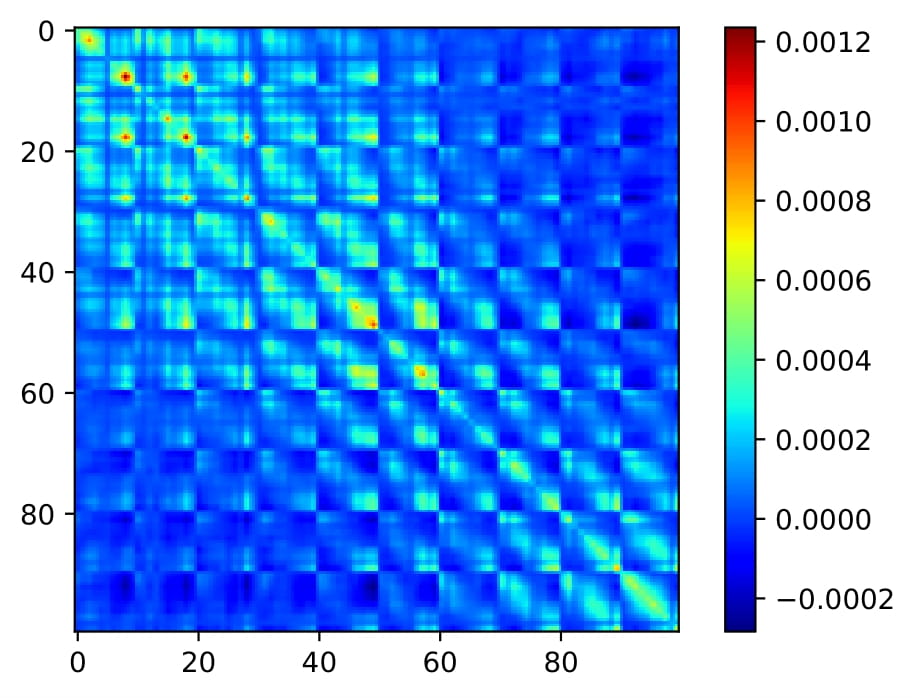}}}
      \subfloat[$\bR_\textrm{D05}$: ex3]{\includegraphics[width=1.8 in]{{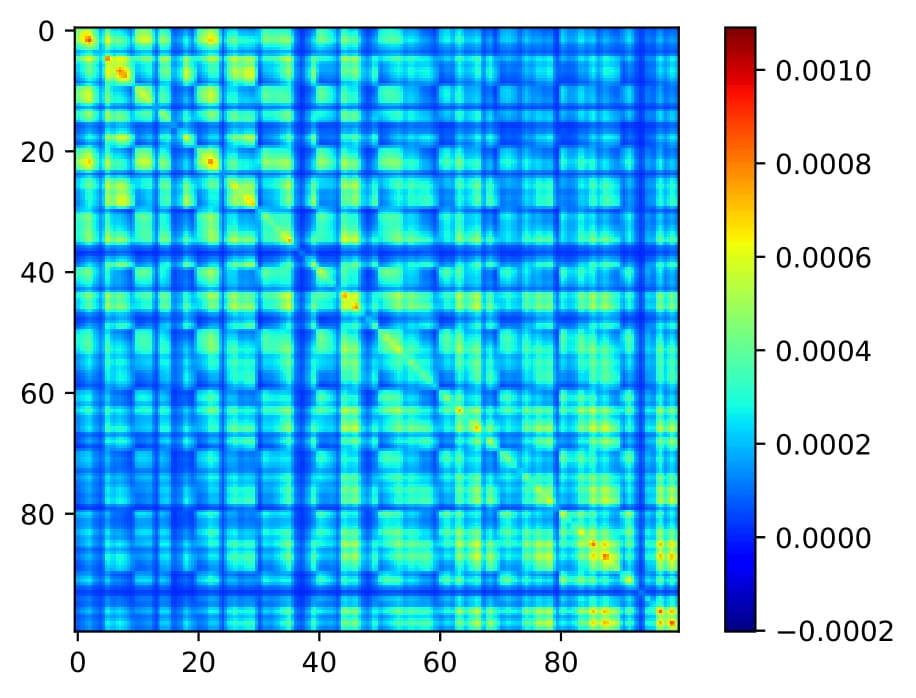}}}
        \subfloat[$\bR_\textrm{D05}$: ex4]{\includegraphics[width=1.8 in]{{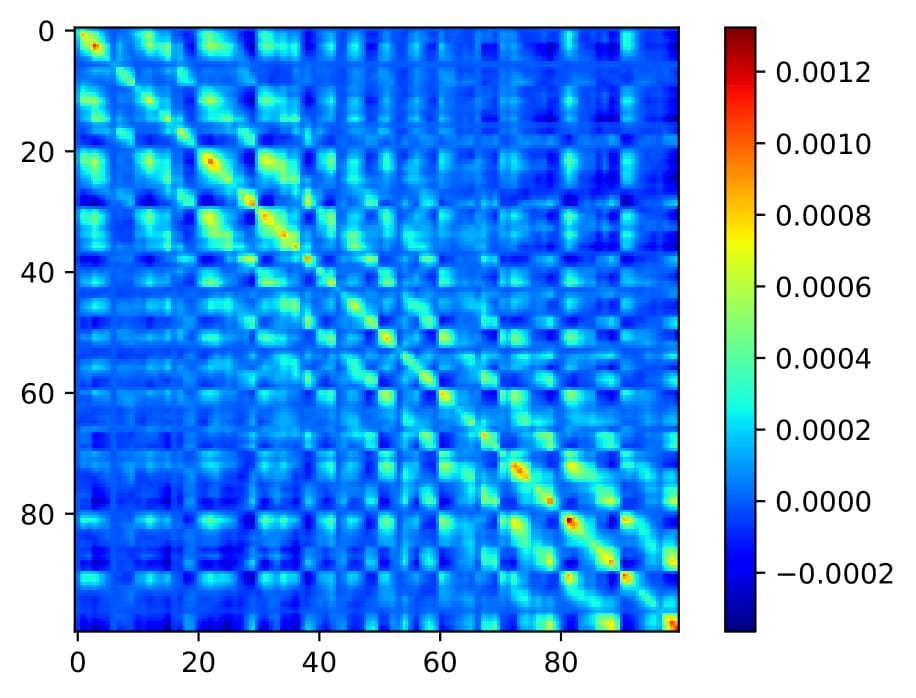}}}
    \caption{Comparison between true and estimated (LSTM1000,D05) $\bR$  matrix which represent the error covariance of 2D observation data, for 4 samples with different $\bD$ and $r$ in the test dataset }
  \label{fig:Rmatrix}
\end{figure}

 Fig.~\ref{fig:1000_1000_lstm_shallowW_evaluation} and Fig. \ref{fig:200_200_lstm_shallowW_evaluation} illustrate respectively the predictions results of LSTM1000 and LSTM200 \sibo{against the true value} on 10000 test examples in the test dataset, which demonstrate that the trained LSTM exhibits a good performance at predicting related values applied to compose $\bR$, including both marginal error variances (i.e., the elements of the $\bD$ vector) and the correlation scale length $r$. The prediction results of LSTM200 (Fig.~\ref{fig:200_200_lstm_shallowW_evaluation}) are almost as accurate as the ones obtained via LSTM1000 (Fig.~\ref{fig:1000_1000_lstm_shallowW_evaluation}). \sibo{Training and evaluating the LSTM network using the first 200 time steps is sufficient to obtain an accurate estimation of the $\bR$ matrix.} The so composed $\bR$ matrices, based on the prediction results from LSTM1000, are shown in Fig.~\ref{fig:Rmatrix} in comparison with the true $\bR$ matrix and that obtained via D05( $q=3$) after regularization. In order to estimate the high dimensional $\bR$ matrix in this application, D05 makes use of the DA residuals every 10 time steps, which is different from the final DA algorithm, in each of the three iterations. Four different sample $\bR$ matrices in the test dataset are displayed in Fig.~\ref{fig:Rmatrix}. We observe that both LSTM and D05 manage to acquire a similar covariance structure as the true $\bR$ matrix while \sibo{the considerable advantage of the LSTM approach can still be noticed}. These results confirm the finding of Fig.~\ref{fig:1000_1000_lstm_shallowW_evaluation} and Fig.~\ref{fig:200_200_lstm_shallowW_evaluation} that \sibo{LSTM can perform well in the parametric prediction of the $\bR$ matrix even in high dimensional systems.}
 
 \setlength{\tabcolsep}{0.8pt}
 
 \begin{table}
\centering
\begin{tabular}{m{2.4cm} m{2.4cm} m{2.4cm} m{2.4cm} m{2.4cm} m{2.4cm} } \toprule
    {parameters} & {True} & {LSTM1000} & {LSTM200}   & {DI01} &  {D05}   \\ \midrule
    
    
    {$\epsilon_{\textrm{mse},(100)}$}  & 0.057013 & 0.056851 & 0.056961 & 0.069590 &  0.072978  \\
    {$\epsilon_{\textrm{mse},(300)}$}  & 0.056051 & 0.056134 & 0.056092 & 0.070572 &  0.073554 \\
    {$\epsilon_{\textrm{mse},(500)}$}  & 0.093322 & 0.093414 & 0.093419 & 0.117149  & 0.110519 \\
    {$\epsilon_{\textrm{mse},(700)}$}  & 0.088271 & 0.088478 & 0.088090 & 0.102763  & 0.102890\\
    {$\epsilon_{\textrm{mse},(799)}$}  & 0.093813 & 0.093404 & 0.093789 & 0.105593  & 0.106912 \\
    {$\bar{\epsilon_{\textrm{mse}}}$}  & 0.117832 & 0.117843 & 0.117785 & 0.127834 & 0.126710  \\
    {Time($s$)}  & \slash  & \textbf{0.26} & \textbf{0.18} &  8.79 &  83.9 
    \\
    \bottomrule
\end{tabular}
\caption{DA performance of the shallow water system evaluated in $\epsilon_{\textrm{mse}}$ based on $\bR$ varied by the predefinition, the algorithm LSTM1000, LSTM200, DI01 and D05. The execution time only consider the process of $\bR$ matrix specification without the DA procedure afterwards}
\label{table:epsion_mse_sw}
\end{table}

To assess the performance of DA accuracy, the metric $\epsilon_{\textrm{mse},(i)}$ (cf., Eq.\ref{eq:mse}) estimated using a set of 53 observation time series, is displayed in Table~\ref{table:epsion_mse_sw}.
Since it would take too much space to display the total ${{\epsilon_{\textrm{mse},(i)}}_{i=0,1,\cdots,800}}$ of all state variables, we select only some of them demonstrated in Table~\ref{table:epsion_mse_sw}, together with the averaged value $\bar{\epsilon_{\textrm{mse}}}$ of all 800 cell coordinates in the field of $u$ and $v$. The displayed results further prove what we have concluded from Fig.~\ref{fig:1000_1000_lstm_shallowW_evaluation} to Fig.~\ref{fig:Rmatrix}, that is, the LSTM-based approaches own an advantage in terms of the DA accuracy compared to D05 ($q=3$) and DI01 ($q=2$) tuning algorithms . Furthermore, the performance of LSTM200 is very close to LSTM1000 with an even slightly smaller average MSE, probably due to the sampling randomness. \sibo{This result further confirms our analysis of Fig.~\ref{fig:200_200_lstm_shallowW_evaluation} that the LSTM model which employs only the first 200 time steps of observation data as input manages to provide an accurate estimation of the $\bR$ matrix.} The averaged computational time (of a laptop CPU) of online covariance tuning/specification is also shown in Table~\ref{table:epsion_mse_sw} where only the evaluation time of the trained LSTM model is taken into account. \sibo{The training of LSTM can be totally performed offline.} As for D05 and D01, we exclude the final DA step (i.e., the computational time is estimated for $q-1$ iterations). As shown in Table~\ref{table:epsion_mse_sw}, the LSTM approach is also considerably faster than traditional tuning methods, which allows a near-real-time application in dynamical systems.

\section{Discussion}
\label{sec:discussion}
 The precision of DA reconstruction/prediction depends heavily on the specification of both the background and the observation error correlation.  The latter is often challenging to estimate in real-world applications because of the dynamic nature of the observation data. Furthermore, the observation matrix $\bR$ can not be empirically estimated from an ensemble of simulated trajectories, unlike the background error covariance. In this paper, we review in detail some well-known observation covariance tuning algorithms \cite{Desroziers01,Descombes2015}, based on time-variant posterior innovation quantities. These methods, being widely adopted in geoscience, rely on 
 some specific prior assumptions such as knowledge of the correlation structure \cite{Desroziers01} or the background matrix \cite{Desroziers2005}. This is difficult to fulfill in some domains where very little knowledge about the prior error is available.\\
 
In this study we have proposed a novel machine learning approach based on LSTM neural networks to predict the $\bR$ matrix using time series observation data as model input. Similar to the work of \cite{Desroziers01} and \cite{Desroziers2005}, $\bR$ is assumed to be time-invariant, at least over a sufficiently long time period. \sibo{Both the Kalman- and variational-type assimilation methods can benefit from the method proposed in this paper for improving the assimilation accuracy.} The proposed data-driven approach also contributes to tackling one of the major bottlenecks of DA: it is time consuming and computationally expensive to update covariance matrix, by mapping raw sensor observations to observation error covariance matrix. In both the Lorenz96 and the shallow water models presented in this paper, the LSTM-based approach displays significant strength, compared to classical posterior tuning methods DI01 and D05, in terms of: (i)estimation accuracy of the observation covariance $\bR$; (ii) reconstruction and prediction accuracy of the DA schema using the estimated $\bR$ matrix; (iii) computational efficiency of the online covariance estimation;  (iv) flexibility of different model parameterization. It is worth mentioning that an important limitation of the proposed LSTM-based method is the specification of $\Phi_\textbf{R}$ which defines the range of parameters for training.\\

\sibo{ Since we assume that the observation matrix is time-invariant, the proposed approach could only deal with fixed sensor placement for dynamical systems, which is also the case of DI01 and D05 tuning algorithms. The possibility of time-variant sensor placement warrants further investigation. As pointed out by \cite{liu2020privacy}, DL model can be stolen
or reverse engineered by model inversion or model extraction attack. Despite the fact that all data used in the current study is generated from toy models, it is important to ensure the data privacy when applying the model to real applications.}
Future research should also consider applying the new method to a broader range of real-world problems, including NWP, hydrology, and object tracking, where the offline data simulation could be more computationally expensive compared to the two test models presented in this paper. To this end, future studies could also investigate the combination of model reduction methods, such as domain localization \cite{cheng2020}, proper orthogonal decomposition, information-based data compression \cite{CHENG2021101405}, auto-encoder neural networks\cite{amendola2020}, and the current covariance estimation method. \sibo{More precisely, the data assimilation can be performed in the compressed low dimensional space (e.g., obtained from POD or auto-encoder). The LSTM-based covariance specification algorithm developed in this work can be used to estimate the observation error covariance matrices in the low dimensional space for improving the accuracy of reduced-order data assimilation approaches.}

\section*{Code availability}
Code for the proposed LSTM-based covariance specification, together with DI01, D05 methods, for both Lorenz and shallow water models is available at\\ https://github.com/scheng1992/LSTM\_Covariance

\section*{Contribution statement}
\noindent S.Cheng: Conceptualization, Methodology, Software, Writing - Original draft preparation \\
M.Qiu: Methodology, Software, Writing - Original draft preparation

\section*{Acknowledgment}
S.Cheng would like to thank Dr. D.Lucor for fruitful discussions about the error covariance computation. This research was supported by EDF R\&D. This research was partially funded by the Leverhulme Centre for Wildfires, Environment and Society through the Leverhulme Trust, grant number RC-2018-023.

\section*{Code availability}
Code for the proposed LSTM-based covariance specification, together with DI01, D05 methods, for both Lorenz and shallow water models is available at\\ https://github.com/scheng1992/LSTM\_Covariance

\section*{Conflict of interest statement}
We declare that we have no financial and personal relationships with other
people or organizations that can inappropriately influence our work, there is no
professional or other personal interest of any nature or kind in any product,
service and/or company that could be construed as influencing the position
presented in, or the review of, the manuscript entitled.

\bibliographystyle{elsarticle-num}
\bibliography{main}
\end{document}